\documentclass[10pt,twocolumn,letterpaper]{article}

\usepackage[pagenumbers]{cvpr} 

\usepackage{graphicx}
\usepackage{amsmath}
\usepackage{amssymb}
\usepackage{booktabs}
\usepackage{times}
\usepackage{epsfig}
\usepackage{graphicx}
\usepackage{amsmath}
\usepackage{amssymb}
\usepackage{pifont}
\usepackage{booktabs}
\usepackage{multirow}
\usepackage{arydshln}
\usepackage{lipsum}  
\usepackage{enumitem}
\usepackage{anyfontsize}
\usepackage[table]{xcolor}
\usepackage{bbm}
\usepackage{arydshln}
\usepackage{pifont}
\usepackage{epsfig}
\usepackage{graphicx}
\usepackage{adjustbox}
\usepackage{array}
\usepackage{booktabs}
\usepackage{colortbl}
\usepackage{float,wrapfig}
\usepackage{hhline}
\usepackage{multirow}
\usepackage{amsmath,amsfonts,amsthm,amssymb}
\usepackage{bm}
\usepackage{nicefrac}
\usepackage{microtype}
\usepackage{enumerate}
\usepackage{changepage}
\usepackage{extramarks}
\usepackage{fancyhdr}
\usepackage{lastpage}
\usepackage{setspace}
\usepackage{soul}
\usepackage{xspace}
\usepackage{booktabs}
\usepackage{longtable}
\usepackage{arydshln}
\usepackage{amssymb}
\usepackage{pifont}
\usepackage{xcolor}
\usepackage[T1]{fontenc}
\usepackage[font=small,labelfont=bf,tableposition=top]{caption}
\DeclareCaptionLabelFormat{andtable}{#1~#2  \& \tablename~\thetable}
\usepackage{capt-of}
 \usepackage{relsize}
\usepackage{epigraph}
\usepackage{diagbox}
\usepackage{makecell}
\usepackage{xr}
\usepackage{xr-hyper}
\usepackage{lipsum}  

\definecolor{darkgreen}{rgb}{0.0, 0.7, 0.33}
\definecolor{darkred}{rgb}{0.6, 0.16, 0.25}
\definecolor{darkblue}{rgb}{0.0, 0.16, 0.75}
\definecolor{maroon}{rgb}{0.76, 0.13, 0.28}
\definecolor{codegreen}{rgb}{0,0.6,0}
\definecolor{codegray}{rgb}{.95,.95, .95}
\definecolor{codepurple}{rgb}{0.58,0,0.82}
\definecolor{backcolour}{rgb}{0.95,0.95,0.92}
\definecolor{p}{rgb}{148,2,209}
\definecolor{darkbrown}{rgb}{0.3, 0.15, 0}
\definecolor{ddarkgreen}{rgb}{0.0, 0.4, 0.23}

\usepackage{mathrsfs}
\usepackage{tabularx}
\newcommand{\evlm}[0]{{$\mathcal{E}$}{-ViLM}\;}
\usepackage[pagebackref,breaklinks,colorlinks]{hyperref}

\usepackage[capitalize]{cleveref}
\crefname{section}{Sec.}{Secs.}
\Crefname{section}{Section}{Sections}
\Crefname{table}{Table}{Tables}
\crefname{table}{Tab.}{Tabs.}

\def\confYear{2024}

\begin{document}

\title{\evlm: Efficient Video-Language Model via Masked Video Modeling \\ with Semantic Vector-Quantized Tokenizer }

\author{\large	Jacob Zhiyuan Fang$^\dag$, Skyler Zheng$^\dag$, Vasu Sharma$^\ddag$, Robinson Piramuthu$^\dag$ \\  \ \ \ \ \  $^\dag$Amazon, \ \ \ \ \ \ \ \ \ \ \ \ \ \ \ \ $^\ddag$Meta  \\
\texttt{\{zyfang, nzhengji,robinpir\}@amazon.com} \qquad  \texttt{vasusharma@meta.com} \\
\\
}

\maketitle

\maketitle

\begin{abstract}
To build scalable models for the challenging real-world tasks, it is important to learn from diverse, multi-modal data in various forms (\eg, videos, text, images). Amongst the existing works, a plethora of them have been focusing on leveraging large but cumbersome cross-modal architectures.
Regardless of their effectiveness, larger architectures unavoidably prevent the models from being extended to real-world applications, so building a lightweight VL architecture and an efficient learning schema is of great practical value.
In this paper, we propose an $\mathcal{E}$fficient \textbf{Vi}deo-\textbf{L}anguage \textbf{M}odel (dubbed as \evlm\!) and a masked video modeling (MVM) schema, assisted with a semantic vector-quantized tokenizer. In particular, our \evlm learns to reconstruct the semantic labels of masked video regions, produced by the pre-trained vector-quantized tokenizer which discretizes the continuous visual signals into labels. We show that with our simple MVM task and regular VL pre-training modelings, our \evlm\!, despite its compactness, is able to learn expressive representations from Video-Language corpus and generalize well to extensive Video-Language tasks including video question answering, text-to-video retrieval, etc. In particular, our \evlm obtains obvious efficiency improvements by reaching competing performances with faster inference speed: \ie, our model reaches $39.3$\% Top-$1$ accuracy on the MSRVTT benchmark, retaining $91.4$\% of the accuracy of state-of-the-art larger VL architecture with only $15\%$ parameters and $94.8\%$ fewer GFLOPs. We also provide extensive ablative studies that validate the effectiveness of our proposed learning schema for \evlm\!.
\end{abstract}

\section{Introduction}

\begin{figure}[t!]
 \vspace{-3mm}
  \begin{center}
    \includegraphics[width=.46\textwidth]{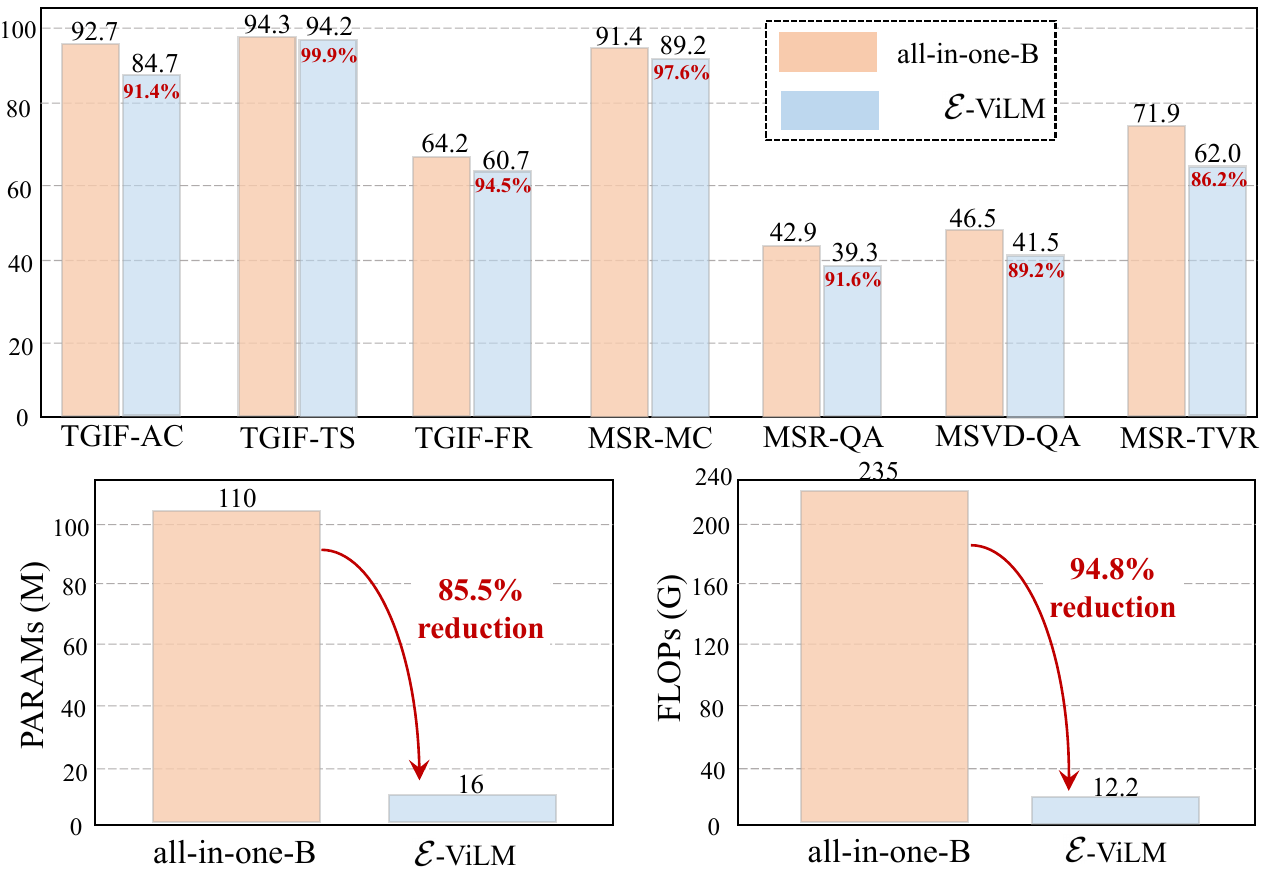} 
  \end{center}
  \vspace{-5mm}
    \caption{\small Comparisons of performances, parameters (PARAMs in M), and inference speed (FLOPs in G) between state-of-the-art video-language model~\cite{jinpeng2022all} as opposed to \evlm\!. We propose a small video-language architecture as well as a novel pre-training schema. \evlm is lightweight: it takes less than $15$\% parameters of all-in-one-B and $94.8\%$ fewer GFLOPs while retaining competing performances on extensive video-language tasks. We show that our model retains prominent performances for  video question answering task on TGIF~\cite{jang2017tgif}, MSRVTT~\cite{xu2016msr}, MSVD~\cite{chen2011collecting}, and text-to-video retrieval task on MSRVTT (R@10).
    }
  \label{fig:abstract}
\end{figure}

The task of video (\textbf{V}) and language (\textbf{L}) pre-training aims to learn joint and robust cross-modal representations from video-text pairs. Recent advancements of VL pre-training have obtained great development and are primarily reflected in the aspects of leveraging more video-text pairs for scaling pre-training~\cite{miech2020end,yuan2021florence,luo2020univl}; superior visual encoder for expressive video representations~\cite{li2020hero,lei2021less,wang2022object}; and unified VL architectures~\cite{fu2021violet,xu2021vlm,wang2021ufo,wang2022git,wang2022all} etc. Despite this, the success of these prior arts unanimously relies on escalating parameters that impede their real-world applications due to the high latency and large memory footprint during inference.

Existing efforts to develop small VL architectures~\cite{wang2020minivlm,fang2021compressing,wang2022efficientvlm,wang2021distilled} or end-to-end cross-modal architectures~\cite{kim2021vilt,wang2022git} for image captioning~\cite{fang2022injecting}, VQA~\cite{cao2022mobivqa}, are restricted to image-text domains only. The challenge then becomes how to train these small VL architectures with increased accuracy. The majority of existing VL pre-training models stand on a common plateau of modeling, i.e., a cross-modal contrastive modeling~\cite{radford2021learning,jia2021scaling} in which visual representations are optimized to be aligned with the matched text while contrasting with the unmatched text; and a masked language modeling~\cite{devlin2018bert} schema in which the model learns to predict randomly masked textual tokens. The modeling of visual representations for VL learning, especially on small VL architectures, are typically less effective and still needs more investigation.

\begin{figure}[t!]
 \vspace{-3mm}
  \begin{center}
    \includegraphics[width=.48\textwidth]{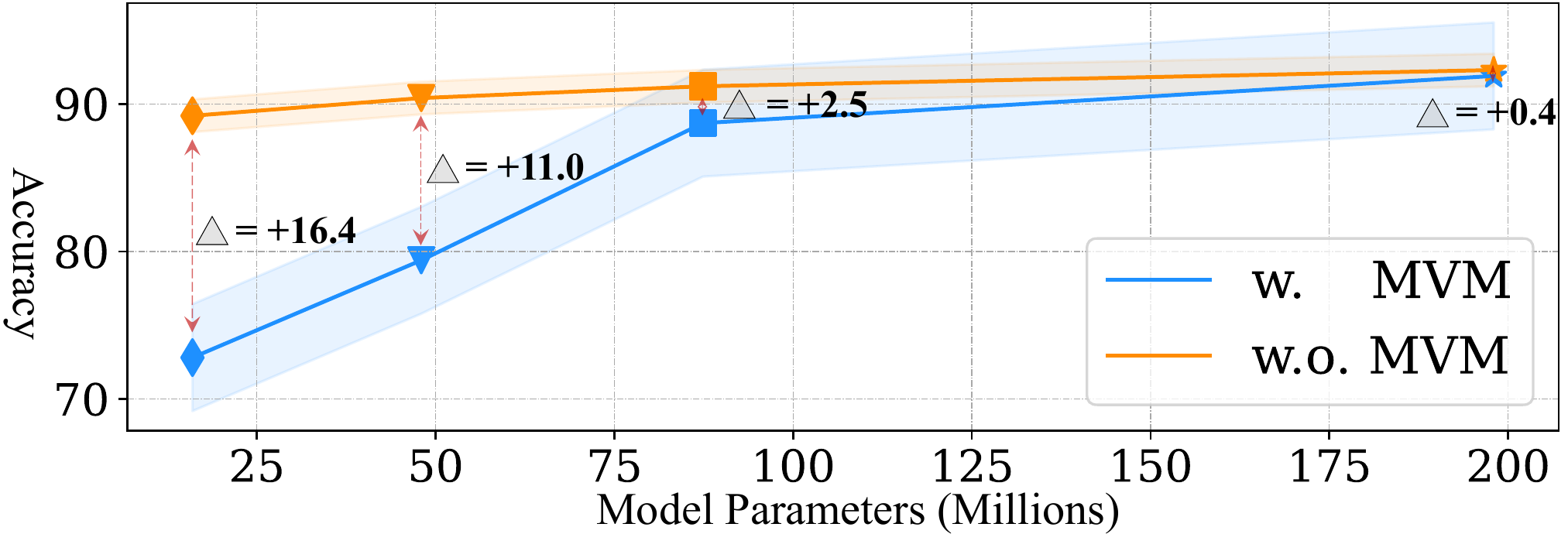} 
  \end{center}
  \vspace{-6mm}
    \caption{\small Masked visual token learning exhibits obvious advantages on smaller VL models, which prefer more discretized visual tokens with highly condensed semantics as supervisions. Baseline denotes the VL models w.o. masked visual token modeling. Experimental details are discussed in Appendix.
    }
  \label{fig:intro}
\end{figure}

Recent studies also witness significant developments in masked token prediction for unsupervised representation learning in both natural language processing (e.g., BERT~\cite{devlin2018bert}, GPT~\cite{radford2018improving,brown2020language, radford2021learning}) and vision domain (e.g., BEiT~\cite{bao2021beit} and MAE~\cite{he2022masked}) from which the learned representations generalize well to diverse tasks.  
Despite remarkable headway made, in the regime of cross-modal tasks dealing with image/video and language, this \emph{``mask-and-predict''} modeling is primarily limited to just the language modality~\cite{su2019vl,li2020unicoder,li2020oscar,zhang2021vinvl,fang2021compressing,wang2020minivlm,dou2022empirical} (reconstructing solely masked language tokens). Hereby, we conjecture that a critical reason that imposes the masked visual token prediction on the VL domain is the inconsistent information density during joint VL pre-training, especially on smaller architectures where visual representations lack expressiveness. In particular: a unified VL architecture (e.g., ViLT~\cite{kim2021vilt} and~\cite{wang2022all}) encodes both modalities where \emph{continuous} visual embedding contains redundant information yet embeddings from language are highly condensed with latent semantics. Such inconsistency prevents the model from inferring contextualized \& aligned cross-modal relations, and it becomes even more visible on smaller VL architectures for highly redundant visual data like videos.  Figure~\ref{fig:intro} shows the Multiple-Choice accuracy of VL models, pre-trained w./w.o Masked Video Token Modeling with different number of learnable parameters: we clearly observe that smaller VL architecture benefits more from discretized masked video token prediction. This clearly indicates that MVM can be the cure of lesion in the furtherance of  a more tiny and powerful VL architecture.

In this paper, we investigate how to set up masked video modeling (MVM) with the goal of obtaining more robust and general VL representations on small VL architectures for diverse video and language tasks. We propose to use a semantic Vector-Quantized (VQ) module to discretize the visual embedding into a set of labels in order to address the aforementioned problem. To accomplish this, we choose the nearest code from a parameterized and learnable codebook as the target tokens for reconstruction. Different from previous efforts like VIOLET~\cite{fu2021violet} and~\cite{fu2022empirical}, we train the video tokenizer without using pixel space, but by reconstructing the words from the video-level captions. 
As a result, the density gap is lessened because the input visuals have been converted into a smaller number of discrete labels. In order to encourage the learning of inferring the latent semantics of masked visual regions from contextualized visual contexts and languages, our proposed MVM is then formulated as predicting the semantic labels of a set of randomly masked video regions. Masked video/language modeling and widely used video-text matching modeling constitute our overall loss. We carry out extensive tests to verify the performance of our proposed MVM. We particularly experimented with VL pre-training on the WebVID~\cite{bain2021frozen} dataset and then assessed the learned representations on a multitude of benchmarks related to video and video-language, including generalization ability, video question answering, and video retrieval by text (linear probe of activity recognition). Our proposed \evlm achieves competing performances consistently across all benchmarks with less than $15$\% learnable parameters and $40$ times fewer GFLOPs than the state-of-the-art (see Figure~\ref{fig:abstract}). The learned representations from the \evlm also exhibit excellent generalization abilities when applied to an activity recognition task.

\noindent To summarize our contribution: 
\begin{itemize} [leftmargin=8pt]
    \setlength\itemsep{0.0em}
    \item We propose a small and novel video-language architecture, \evlm\!, which achieves prominent accuracy \& inference speed trade-off. To the best of our knowledge, this is the very initial attempt that leverages a small video-language model for VL representation learning.
    \item To counter the pain point of training small VL architecture caused by inconsistent cross-modal information density, we propose a novel semantic reconstruction based video tokenizer for discretized masked video modeling. This facilitates our small VL model to efficiently learn from large-scale video-text pairs for expressive representations.
    \item Extensive quantitative experiments validate the effectiveness of our proposed model and learning schema. Our proposed \evlm achieves competing results across extensive benchmarks with obvious inference advantages.
\end{itemize}

\section{Related Work}

\begin{figure*}[t]
  \begin{center}
    \includegraphics[width=.96\textwidth]{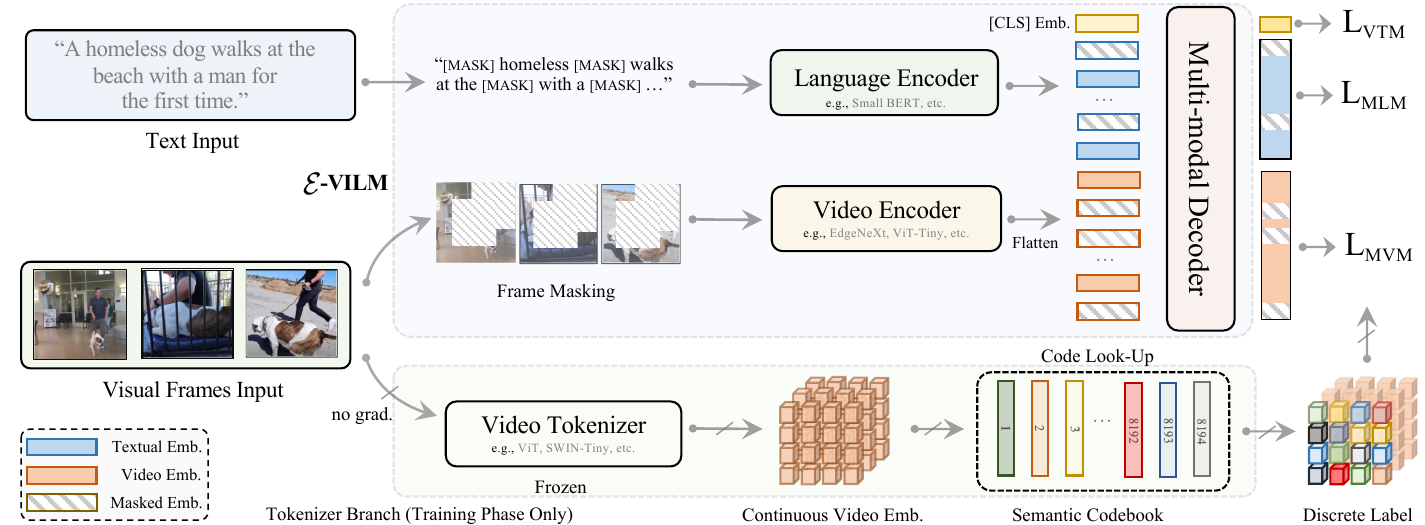} 
  \end{center}
  \vspace{-3mm}
    \caption{\small \textbf{Overall training pipeline of our proposed \evlm}. We leverage efficient design in pursuit of a lightweight Video-Language architecture with fast inference speed. To achieve this, we utilize EdgeNext~\cite{maaz2022edgenext} as the video backbone that encodes the masked video frames as video embedding, and a tiny variation of BERT~\cite{devlin2018bert}, as the language encoder. A multi-modal decoder module is composed of a stack of self-attention blocks~\cite{vaswani2017attention} with decreased hidden dimensions for video/language embedding fusion. For competing performances, we propose to employ a video tokenizer (composed of an encoder with ViT~\cite{radford2021learning} or Video SWIN~\cite{liu2022video,liu2021swin}, and a vector-quantized tokenizer) to map the \textit{unmasked} frames into continuous video embedding, which are later discretized into semantic tokens by nearest neighbor look-up. Our masked video modeling ($\mathcal{L}_\text{MVM}$) is formulated as predicting the categorical labels of the masked video embedding. Our overall training schema also includes the video-text matching modeling ($\mathcal{L}_\text{VTM}$) and masked language modeling ($\mathcal{L}_\text{MLM}$). Note that the tokenizer branch (in the green rectangle) is included only during the training phase providing semantic labels. (Best viewed in color).
    }
  \label{fig:architecture}
\end{figure*}

\vspace{1mm}
\noindent \textbf{Efficient Vision-Language Models.} The remarkable performance of most VL models~\cite{mu2022slip,goel2022cyclip,huo2021wenlan,flava,perceiver,perceiverIO,gao2022transform,lei2021less,yang2021just,xu2021videoclip,sun2019videobert,xu2021videoclip} is closely dependent on the great visual representations from expressive visual encoders like object detector~\cite{anderson2018bottom, li2020oscar}. Zhang et al.~\cite{zhang2021vinvl} develop improved Faster-RCNN~\cite{ren2015faster} for the sake of better visual representations that substantially uplifts the results on various VL tasks. Despite so, cumbersome visual encoder like object decoder arguably hinders the usefulness of VL models in real-world applications due to their demands of high computational resources and are de facto the major computational burden in VL architecture. To resolve this challenge, many recent efforts propose to optimize the detector~\cite{wang2020minivlm,fang2021compressing}, utilize grid visual representations~\cite{jiang2020defense,kim2021vilt,fang2022injecting,dou2022empirical,wang2021ufo} with no regional operations in the detector.  There also emerge many efforts to build efficient Vision-Language models. For instance,~\cite{wang2020minivlm} proposes to utilize an EfficientNet variations~\cite{tan2019efficientnet} as the detector for visual feature extraction, which largely minimizes the inference bottleneck. Other follow-ups exploit knowledge distillation technique~\cite{fang2021compressing} to improve the accuracy of VL models.~\cite{jiang2020defense,luo2020univl,kim2021vilt,fang2021compressing} explore the feasibility of using end-to-end models to reduce the inference time on object detector than two-stage VL architectures.

\noindent \textbf{Masked Auto-encoder.}
Recent studies also witness significant development in masked token prediction for unsupervised representation learning in both natural language processing, e.g., BERT~\cite{devlin2018bert} and GPT~\cite{radford2018improving,brown2020language, radford2021learning}. 
This \textit{mask-and-predict} pre-training manner extends to the visual representation learning: BEiT~\cite{bao2021beit} for the first time proposes to tokenize the continuous visual embedding into discrete tokens and then reconstruct the randomly masked token. MAE~\cite{he2022masked} further simplifies this by directly predicting the masked patches and using their pixel values as the reconstruction target. Follow-up works further advance this to different architectures~\cite{li2022uniform}, various visual data (\ie, videos~\cite{feichtenhofer2022masked,min2022voxel,tong2022videomae}, depth image~\cite{yan2022multi} etc). In particular, concurrent works from Tong et al.~\cite{videoMAE} and Feichtenhofer et al.~\cite{videoMAE2} both show that video data, due to its high spatio-temporal redundancy, requires a significantly high masking ratio. 
Chen et al.~\cite{chen2019uniter} adopts a masked region modeling to reconstruct the masked regional representations of the image for VL pre-training but with limited improvement. 
Bachmann et al.~\cite{mtmae} also~\cite{geng2022multimodal} further extend this to a multi-modal-multi task setup where they show that additional modalities can be incorporated into the training objective by multi-modal masking.
In particular, VIOLET~\cite{fu2021violet} also proposes to leverage the video tokenizer to provide discretized video patch labels for masked video modeling. Our work differentiates with~\cite{fu2021violet} majorly from two perspectives: 1. we focus on developing small and efficient architectures. 2.~\cite{fu2021violet} leverages tokenizer of DALL-E~\cite{ramesh2022hierarchical} as video tagger, which is obtained from a pixel-level re-construction task while ours is trained from semantic-level re-construction.

 \begin{figure*}[t]
  \begin{center}
    \includegraphics[width=1.\textwidth]{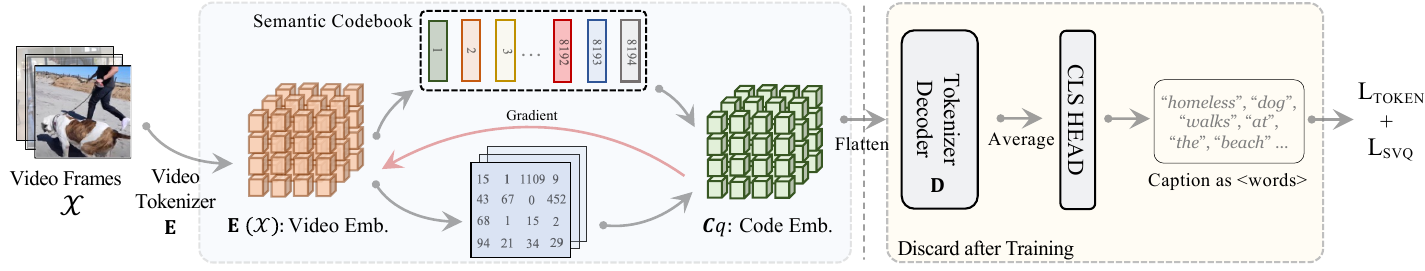} 
  \end{center}
  \vspace{-4mm}
  \caption{\small \textbf{Overall training pipeline of our proposed video tokenizer}. Our tokenizer is an Encoder-Decoder architecture where video frames are encoded into continuous video embedding, and discretized into code via nearest neighbor search. We optimize the tokenizer by reconstructing the words from captions via a multi-label classification task. 
    }
  \label{fig:tokenizer}
\end{figure*}

\section{Efficient Video-Language Modeling}
In this section, we describe how we build and train our efficient Video-Language Model (\evlm) that is smaller and faster. This section first discusses in detail about the architectural design of the model, followed by our proposed Masked Video Modeling aiming for more efficient VL pre-training and ends with our overall training schema.

\subsection{Model Architecture}
Figure~\ref{fig:architecture} illustrates the overall architecture of our \evlm\!, which is an end-to-end architecture consisting of : a Language Encoder ($\mathbf{ENC}_\text{L}$) that encodes the text ($\mathcal{T}$) into a sequence of continuous embedding, Video Encoder ($\mathbf{ENC}_\text{V}$) that encodes stack of sparsely sampled video frames ($\mathcal{I}$), and a transformer decoder ($\mathbf{DEC}_\text{VL}$) that allows cross-modal interaction across Visual and Language modalities:
\begin{equation}
\footnotesize{
    \mathbf{e}^{V} = \mathbf{ENC}_\text{V}(\mathcal{I}), \ \  \mathbf{e}^{L} = \mathbf{ENC}_\text{L}(\mathcal{T}), \ \  \mathbf{h} = \mathbf{DEC}_\text{VL}(\mathbf{e}^{V}, \mathbf{e}^{L})},
\end{equation}
where $\mathbf{e}^{V/L}$ denotes the visual/language embedding. In quest of a lightweight VL model, we purposefully constrain the size of each module in \evlm by leveraging a few efficient but well-performing deep architectures across both Vision and Language benchmarks. 

\vspace{2mm}
\noindent\textbf{Video Encoder.}  In our work, we resort to a lightweight 2D visual architecture as the visual encoder to obtain frame-wise grid representations as visual embedding. To be specific, given the sparsely sampled $T$ video frames $\mathcal{I} = \{\mathcal{I}_0 \dots \mathcal{I}_T\}$, $\mathcal{I}_i \in \mathbb{R}^{H\times W\times 3}$, each frame is represented as continuous embedding $\mathbf{e}^V_i \in \mathbb{R}^{H/S\times W/S\times D}$, where $S$ and $D$ denote the down-sample ratio of the network and the dimension of embedding, respectively. We then flatten and concatenate all embeddings, followed by a linear transformation for dimension adjustment to produce the video-level visual representation $\mathbf{e}^V$.

\vspace{2mm}
\noindent\textbf{Language Encoder.} Transformer architecture~\cite{vaswani2017attention} and its variations~\cite{devlin2018bert,radford2018improving} achieve exceptional performance on multiple natural language understanding tasks. Yet, transformer like architectures suffer from slow inference speeds primarily because of the quadratic complexity of the self-attention mechanism and a large numbers of parameters. Although improved works~\cite{tsai2019small,shen2021efficient} take advantage of efficient/local self-attention mechanism or knowledge distillation~\cite{jiao2019tinybert} technique to address this, their performances still fall short of expectation. Instead, we propose to utilize a small BERT architecture with decreased hidden dimension and squeezed vocabulary where only frequent words are retained in favor of a more tiny tokenizer layer. A linear projection layer maps the language embedding to a consistent dimension at the end. We introduce detailed architectural settings in the experiment section.

\vspace{2mm}
\noindent\textbf{Multi-modal Decoder.} Our multi-modal decoder is composed of a stack of transformer blocks that takes  the concatenation of visual and language embeddings as input, enabling the cross-modal interaction. To incorporate information for distinguishing between the two modalities, we add token type embeddings $\mathbf{t}^L$ and $\mathbf{t}^V$ to language embedding and video embedding, respectively. We also attach a special classification token \texttt{[CLS]} (as shown in Figure~\ref{fig:architecture}):
\begin{equation}
    [\mathbf{h}^\text{CLS}, \mathbf{h}^\text{L}, \mathbf{h}^V] = \mathbf{DEC}_\text{VL}([\mathbf{e}^\text{CLS}, \mathbf{e}^L+\mathbf{t}^L , \mathbf{e}^V+\mathbf{t}^V]).
\end{equation}
Similar to the language encoder, we decrease the hidden dimension in each block.

\subsection{MVM with Semantic Video Tokenizer}
We propose a novel masked video modeling (MVM) based on a semantic video tokenizer to assist VL representation learning. Masked language modeling (MLM) obtains great success~\cite{devlin2018bert} in a series of natural language comprehension tasks. Recent efforts on visual masked auto-encoder~\cite{he2022masked,tong2022videomae} also show prominent effectiveness on self-supervised representation learning.
In essence, the visual model learns to restore the masked raw pixels based on the observable pixels, which are continuous and high-dimensional, while the prediction target in MLM is the language token with high semantics and dense information. In particular, VL architecture encodes both modalities where continuous visual embedding contains redundant information yet semantics from language are highly condensed. Such inconsistency prevents the model from inferring out contextualized \& aligned cross-modal relations, which play an essential role in VL pre-training~\cite{cao2020behind}, especially on highly redundant visual data like videos. For this reason, we propose to leverage an independent Tokenizer branch (described in a later section) as a video tagger, providing one-hot semantic labels $\mathbf{y}_i^V$ for the $i$-th masked video region. 
To be specific, we randomly mask out video frame patches by replacing them with zero-value pixels. Our proposed MVM aims to reconstruct the semantic labels of the masked regions via a classification task. And we use cross-entropy to measure this error:
\begin{equation}
\mathcal{L}_\text{MVM} = - \sum_{i=1}^{N} \frac{1}{N} \text{CE} (\hat{\mathbf{y}_i}^V, \mathbf{y}_i^V),
\end{equation}
where $N$ represents the number of masked regions and $\hat{\mathbf{y}_i}^V$ is the predicted semantic probability of the $i$-th masked video region, obtained by a semantic classification head: $\hat{\mathbf{y}_i}^V = \mathbf{HEAD}^{V}(\mathbf{h}_i^V)$. In the later section, we explain how we construct our video tokenizer and discuss how it differentiates from previous efforts.

\subsection{Semantic Vector-Quantized Video Tokenizer}
A recent work leverages a visual tokenizer to discretize the visual regions for self-supervised representation learning~\cite{peng2022beit}. \cite{lu2022unified,fu2021violet} further extend this manner to Vision-Language learning and utilize a pre-trained visual tokenizer (i.e., tokenizer of DALL-E~\cite{ramesh2021zero} and VQ-GAN~\cite{esser2021taming}) as a tagger. However, a notable defect of such a tokenizer comes from its lack of semantics, as they are unanimously obtained from the pre-task of reconstructing raw image pixels in an auto-encoder fashion. To overcome this, we propose to construct our video tokenizer by reconstructing the semantics of the videos.

Following~\cite{van2017neural,esser2021taming}, we first learn an asymmetric Encoder-Decoder model consisting of a ViT~\cite{radford2021learning} or VideoSWIN~\cite{liu2022video} based encoder ($\mathbf{E}$) and
a decoder ($\mathbf{D}$) that is composed of stacks of self-attention blocks. 
It learns to represent the video frames with codes from a learnable and discrete codebook with $M$ latent code embedding $\mathcal{C} = \{ \mathbf{c}_1 \dots \mathbf{c}_M \}$, with $\mathbf{c}_j \in \mathbb{R}^{D}$ (see Figure~\ref{fig:architecture}). Our vector-quantization module $\mathbf{q}(\cdot)$ approximates a set of latent codes $\mathcal{C}_q$ for input video frames by a nearest neighbor look-up using the shared embedding space such that: 
\begin{equation}
    \mathcal{C}_q = \mathbf{q}\Big(\mathbf{E}(\mathcal{X})\Big) := \bigg(\underset{\mathbf{c}_j \in \mathcal{C},\mathbf{x}_i \in \mathcal{X}}{\text{argmin}}\Big|\big| \mathbf{E}(\mathbf{x}_i) - \mathbf{c}_j \big|\Big|\bigg),
\end{equation}
with $\mathcal{X} = \{\mathbf{x}_i\}_{i=1}^{T\times H/h \times W/w}$, and $\mathbf{x}_i \in \mathbb{R}^{h\times w\times 3}$ denotes $i$-th the image patch obtained from $T$ sampled image frames via non-overlapping partitioning.

We introduce the semantic reconstruction task by predicting the discrete words $\mathcal{T} = \{\mathbf{w}_k \}$ of the video caption with latent codes by the decoder. We compute the multi-label probability using a classification head ($\mathbf{HEAD}^S$) given the averaged output embedding from the decoder ($\mathbf{D}$):
\begin{equation}
    \hat{\mathbf{y}}^{S} = \mathbf{HEAD}^S\Big(\frac{1}{|\mathcal{X}|}\sum\mathbf{D}(\mathcal{C}_q)\Big).
\end{equation}
Considering the extremely imbalanced word distribution and the rarity of positive samples, we adopt focal loss for tokenizer training~\cite{ben2020asymmetric,fang2021compressing}, which shows outstanding results on a multi-label classification task:
\begin{equation}
      \mathcal{L}_\text{TOKEN} = \frac{1}{K} \sum^{K}_{k=1}
\begin{cases}
    \ \ \ \ \ \ \ \  \text{log}(\hat{\mathbf{y}_k}^S), & \mathbf{y}_k^S=1, \\ \hat{\mathbf{y}_{k}}^S \cdot \text{log}(1-\hat{\mathbf{y}_k}^S), & \mathbf{y}_k^S=0,
\end{cases} 
\end{equation}
where $\hat{\mathbf{y}_k}^S \in [0,1]$ is the predicted probability for the $k$-th category, and $K$ denotes the size of the vocabulary. Intuitively, our tokenizer is trained to represent visual frames as a distribution over latent constituents, and these latent codes contain vivid semantics that are essential to Vision-Language pre-training task. Because back-propagation is not possible due to the non-differentiable  quantization step, we copy the gradients over $\mathcal{C}_q$ directly to the $\mathbf{E}(\mathcal{X})$ for encoder (straight-through gradient estimating~\cite{bengio2013estimating}). Our Semantic Vector-Quantized Tokenizer is optimized by the objective:
\begin{equation}
\small
    \mathcal{L}_\text{SVQ} = \mathcal{L}_\text{TOKEN} + \big||sg[\mathbf{E}(\mathcal{X})] - \mathcal{C}_q\big||^2_2  + |\big|\mathbf{E}(\mathcal{X}) - sg[\mathcal{C}_q]|\big|^2_2,
 \end{equation}
and $sg[\cdot]$ denotes the stop-gradient operation~\cite{van2017neural}, which ensures the encoder commits to an embedding and its output does not grow arbitrarily. After the training, we retain only the encoder and the vector-quantized module as our tokenizer. Note that during VL pre-training, the tokenizer is capable of predicting the discrete semantic labels either online or off-the-shelf. 

\subsection{Video-Language Pre-Training}
The training of \evlm consists of two stages, where we first obtain the video tokenizer by optimizing it jointly with $\mathcal{L}_\text{SVQ}$ and $\mathcal{L}_\text{TOKEN}$. 
For the overall Video-Language pre-training, we use the common VL pre-training modeling objectives: masked language modeling loss ($\mathcal{L}_\text{MLM}$), where a certain ratio of the textual words are masked and the model predicts their label; also
Video-Text matching loss ($\mathcal{L}_\text{VTM}$) where
the model is expected to predict whether the video-text
pair matches. Our overall loss at the pre-training stage is the combination of the aforementioned items:
\begin{equation}
    \mathcal{L} = \mathcal{L}_\text{MLM} + \mathcal{L}_\text{VTM} + \mathcal{L}_\text{MVM}.
\end{equation}

\begin{table*}[t]
\centering
    \setlength\tabcolsep{4.1pt}
    \begin{tabular}{lccccp{-4mm}cccp{1mm}ccp{1mm}c}
        \toprule
        \multirow{2}{*}{\textbf{Methods}} & \multirow{2}{*}{\textbf{\# P}} & \multirow{2}{*}{\textbf{FP}} & \multirow{2}{*}{\textbf{\# F}}& \multirow{2}{*}{\textbf{P.T.}}& & \multicolumn{3}{c}{\texttt{\textbf{TGIF}}} &  & \multicolumn{2}{c}{\texttt{\textbf{MSRVTT}}} &  & \texttt{\textbf{MSVD}} \\
        \cmidrule{7-9} \cmidrule{11-12} \cmidrule{14-14} 
       & &  &  &  &  & Action & Transition & Frame &  & MC & QA &  & QA \\
        \midrule
        SSML~\cite{amrani2021noise}  & -  & - & - & HowTo100M & & - & - & - &  & - & $35.0$ &  & $35.1$ \\
        CoMVT~\cite{seo2021look} & -  & - & $30$  & HowTo100M & & - & - & - &  & - & $39.5$ &  & $42.6$ \\ 
        ClipBERT~\cite{lei2021less} & $137$M  & $3.8$K & $16$ & {\fontsize{8}{8}\selectfont COCO~\cite{lin2014microsoft} + VG~\cite{krishna2017visual}} & & $82.8$ & $87.8$ & $60.3$ &  & $88.2$ & $37.4$  &  & - \\
        JustAsk~\cite{yang2021just} & $157$M  & $4.4$K & - & {\fontsize{8}{8}\selectfont HowToVQA69M~\cite{yang2021just}} & & - & - & - &  & - & $41.5$  &  & $46.3$ \\
        all-in-one-S~\cite{jinpeng2022all} & $33$M  & $134$ & $3$ & {\fontsize{8}{8}\selectfont Web2.5M+HT100M} & & $91.2$ & $92.7$ & $64.0$ &  & - & $39.5$ & & $41.7$ \\ 
        all-in-one-B~\cite{jinpeng2022all} & $110$M  & $235$ & $3$ & {\fontsize{8}{8}\selectfont Web2.5M+HT100M} & & $\textbf{92.7}$ &  $94.3$ & $64.2$ &   & $91.4$ & $42.9$ & &$46.5$\\ 
        VIOLET~\cite{fu2021violet} & $198$M  & $4.9$K & $5$ & {\fontsize{9}{9}\selectfont Web2.5M+CC} & &  $92.5$ & $\textbf{95.7}$ & $\textbf{68.9}$ &  & $\textbf{91.9}$ & $\textbf{43.9}$ &  & $\textbf{47.9}$ \\
        all-in-one-T~\cite{jinpeng2022all} & $12$M   & $18.1$ & $3$ & {\fontsize{8}{8}\selectfont Web2.5M+HT100M} & & $80.6$ & $83.5$ & $53.9$  &  &  \ \ $64.8^*$ & \ \ $36.8^*$ & & \ \ $32.2^*$ \\ 
        \evlm  & $16$M  & $12.2$ & $4$ & Web2.5M & & ${\textbf{84.7}}$ & ${\textbf{94.2}}$ & ${\textbf{60.7}}$ & & ${\textbf{89.2}}$ & ${\textbf{39.3}}$ & & ${\textbf{41.5}}$ \\
        \bottomrule
    \end{tabular}
    \vspace{2mm}
    \caption{Comparison of \evlm with existing methods with VL pre-training on video question answering and multiple-choice (MC) tasks. \textbf{P.T.} denotes the pre-trained data. \# \textbf{P} and \# \textbf{F} represent the number of parameters and number of frames. The bottom part of the table shows the results of small VL architectures ( $\leq 20$M parameters). $^*$ denotes the results are re-produced using official released pre-trained checkpoint\cite{jinpeng2022all}. \textbf{FP} denotes the FLOPS of models (in G). Best results are \underline{underlined}. }
    \vspace{-2mm}
    \label{table:vqa}
\end{table*}

\section{Experiments}
We now introduce the detailed implementation of \evlm and also discuss the empirical results that validate the effectiveness of our proposed training schema. To highlight the generalizability of \evlm\!, we benchmark performances of \evlm and compare them with prior arts on multiple video-language benchmarks. We then empirically study the benefits of different VL models and conduct ablative studies for various settings. To this end, we discuss in-depth about the efficiency of \evlm and outline potential future works of efficient Vision-Language learning.

\subsection{Datasets}

\vspace{2mm} 
\noindent\textbf{Pre-training Dataset.} In our experiment, we pre-train our model on {WebVid-2.5M}~\cite{bain2021frozen}, which contains $2.5$M video-text pairs scraped from the web with captions describing the holistic semantics of the video. Videos from WebVid are averagely around $18$ seconds with ~$20$ words per caption.

\vspace{2mm} 
\noindent\textbf{Downstream Datasets.}
To assess the representations learned by \evlm\!, we conduct extensive evaluations across multiple video-language downstream tasks with more than $10$ benchmarks: spanning over video question answering task on MSRVTT~\cite{xu2016msr}, MSVD~\cite{chen2011collecting}, TGIF-QA (including  TGIF-Action, TGIF-Transition and TGIF-Frame)~\cite{li2016tgif,jang2017tgif} datasets; video retrieval by text task on MSRVTT, DiDeMo~\cite{anne2017localizing}. Meanwhile, in order to further evaluate the generalizability of \evlm\!, we continue to transfer the learned VL representations for linear-probe action recognition task on HMDB51~\cite{kuehne2011hmdb} and Kinetics-400~\cite{kay2017kinetics} datasets, as well as the zero-shot multiple-choice task on MSRVTT. We report the performance of Recall at $K$ ($R@K, K=1, 5, 10$) for text-to-video retrieval task; and report the accuracy on VQA task, action recognition and zero-shot multiple choice tasks.

\vspace{2mm} \noindent\textbf{Architectural Details.}
We build our \evlm based on an efficient visual architecture, EdgeNext~\cite{maaz2022edgenext}, that involves both convolution layers and efficient local attention operation but with a light computational burden. Specifically, our visual encoder (EdgeNext\_small) only consumes $1.26$G FLOPs for one $256 \times 256$ frame with $5.59$M parameters. Note that our \evlm is not architecture-specific and we choose EdgeNext~\cite{maaz2022edgenext} in our implementation due to their well-organized open-source code and good accuracy-speed trade-off. For the language encoder, we adopt a tiny BERT variation with decreased hidden size ($128$) and fewer attention heads per layer ($4$). In order to further compress the language branch, we truncate the vocabulary of the tokenizer by retaining only the top $30\%$ most frequent words in COCO dataset~\cite{lin2014microsoft} to avoid dataset bias. This additionally squeezes the language encoder with only $2.4$M learnable parameters. 
Our cross-modal decoder is composed of $4$ transformer blocks with decreased hidden size ($384$). In total, our \evlm accommodates $16$M parameters (without a task specific head) and can encode video (4 frames)-caption pairs with $12.2$ FLOPs(G), which is obviously advantageous compared with previous state-of-the-art. Table~\ref{table:vqa} also provides the specific number of parameters and FLOPs comparisons with other VL architectures in regard to efficiency.

\begin{table*}[t!]
	\hfill
	\begin{minipage}{0.68\linewidth}
    \centering
    \setlength{\tabcolsep}{3.8pt} 
    {    
    \begin{tabular}{lcccccc}
    \toprule
    \multirow{2}{*}{\textbf{Methods}}  & \multirow{2}{*}{\textbf{P.T.}} & & \multicolumn{1}{c}{\texttt{\textbf{MSRVTT}}} & & \multicolumn{1}{c}{\texttt{\textbf{DiDeMo}}}  \\
    \cmidrule{4-4} \cmidrule{6-6}
    &  & & text-to-video & & text-to-video  \\
    \hline
    HT100M~\cite{miech2019howto100m} & HT100M & &  $14.9$ / $40.2$ / $52.8$  & & -  \\
    {HERO~\cite{li2020hero}} & HT100M & & $16.8$ / $43.4$ / $57.7$  & & \ \ \ $2.1$ / $11.4$ / $36.1$ \\
    ClipBERT~\cite{lei2021less} & {\fontsize{8}{8}\selectfont COCO~\cite{lin2014microsoft} + VG~\cite{krishna2017visual}} & &  $22.0$ / $46.8$ / $59.9$  & & $20.4$ / $48.0$ / $60.8$ \\
    TACo~\cite{yang2021taco} & HT100M & & $28.4$ / $57.8$ / $71.2$ &  & - \\
    VideoCLIP~\cite{xu2021videoclip} & HT100M &  & $30.9$ / $55.4$ / $66.8$ & & $16.6$ / $48.0$ / \ \ \ - \ \ \    \\
    Frozen~\cite{bain2021frozen} & {\fontsize{8}{8}\selectfont CC~\cite{sharma2018conceptual} + Web2.5M} & & $32.5$ / $61.5$ / $71.2$ & & $31.0$ / $59.8$ / $72.4$  \\
    VIOLET~\cite{fu2021violet} & {\fontsize{8}{8}\selectfont CC~\cite{sharma2018conceptual} + \newline Web2.5M} & & $34.2$ / $63.5$ / $73.6$ & & $32.9$ / $63.0$ / $74.7$  \\
    all-in-one-B~\cite{wang2022all} &  {\fontsize{8}{8}\selectfont HT100M + \newline Web2.5M} & & \textbf{$39.5$ / $63.3$ / $71.9$}  & & - \\ 
    \hline
    all-in-one-T~\cite{wang2022all} & {\fontsize{8}{8}\selectfont HT100M + \newline Web2.5M} & & \ $16.3$ / $37.4$ / $53.1^*$ & & \ \  $12.5$ / $26.4$ / $38.9^*$ \\ 
    \evlm & Web2.5M & & \textbf{$27.0$ / $51.7$ / $64.4$} &  & \textbf{$23.8$ / $44.9$ / $53
.2$}  \\
    \bottomrule
    \end{tabular}}
    \vspace{1mm}
    \caption{Comparison of \evlm with state-of-the-art VL models on 
    \textbf{text-to-video-retrieval} task. Results are reported on R@1 / R@5 / R@10. \textbf{P.T} denotes the VL pre-training corpus.}
    \label{table:retrieval}
	\end{minipage} \hfill
	\begin{minipage}{0.28\linewidth}
    \centering
    \renewcommand{\arraystretch}{.85} 
    \setlength\extrarowheight{3pt}
    \scalebox{0.9}
    { 
    \small
    \begin{tabular}{lcp{1mm}c}
    \toprule
    \multirow{2}{*}{\textbf{Methods}}  & \multicolumn{1}{c}{\texttt{\textbf{HMDB51}}}  & & \multicolumn{1}{c}{\texttt{\textbf{K400}}} \\
    \cmidrule{2-2} \cmidrule{4-4}
    & {Top-1} & & {Top-1} \\
    \hline 
    Frozen~\cite{bain2021frozen} & $54.3$ & & $50.5$\\
    T-AVG.~\cite{wang2022all} & $43.1$ &  & $44.3$\\
    all-in-one-B & $52.9$ & & $50.8$\\ 
    \hline
    all-in-one-T & \ \ $18.4^*$ & & \ \ $16.7^*$ \\ 
    \evlm  &  $39.6$  & & $39.7$\\
    \evlm$^{\text{MVM}}$  &  $41.6$  & & $41.9$\\
    \bottomrule
    \end{tabular}
    }
    \vspace{2mm}
    \caption{Activity recognition via linear probe on Kinetics-400~\cite{kay2017kinetics} and HMDB51~\cite{kuehne2011hmdb} datasets.}
    \label{table:activity}
	\end{minipage} 
	\vspace{-4mm}
\end{table*}

We instantiate our video tokenizer with different visual backbones and compare their performances: namely, a ViT~\cite{dosovitskiy2020image} architecture pre-trained by~\cite{radford2021learning}. Our tokenizer also includes a decoder which consists of $8$ layers of transformer architecture, yet this module is dropped after the training phase of the video tokenizer. Our semantic codebook contains $9,420$ learnable prototype embedding in the dimension of $32$ initialized randomly. To compare the effects of various tokenizers, we also explore to utilize other architectures as the tokenizer, \eg, discrete variational auto-encoder (dVAE) from DALL-E~\cite{ramesh2021zero} which is trained by reconstructing the RGB pixels, and a Video Swin Transformer (VideoSwin-Tiny)~\cite{liu2022video} pre-trained on Kinetics-400~\cite{kay2017kinetics} using our learning schema.
Our video tokenizer ``\textit{escorts}'' the pre-training of \evlm in the evaluation mode, so it barely takes up resources during VL pre-training.

\vspace{2mm} \noindent\textbf{Training.} For VL pre-training, we sparsely sample $4$ video frames per video and resize them into $256 \times 256$ after random augmentation. Different masking strategies and masking ratios, as previous works indicate~\cite{videoMAE,tong2022videomae,fu2022empirical,fu2021violet}, may affect the representation learning. We adopt block-wise masking~\cite{bao2021beit} with masking ratio $40\%$. We use optimizer adamW~\cite{loshchilov2017decoupled} with an initial learning rate of $1e-4$, weight decay of $0.01$ during pre-training and train \evlm on WebViD for $20$ epochs with $\mathcal{L}_\text{MVM}+\mathcal{L}_\text{VTM}+\mathcal{L}_\text{MLM}$ jointly. For MLM, we randomly mask $15\%$ of the words during training.
For downstream tasks, we stick to the sparsely sampled $4$ frames and use task-specific learning rates/epochs. Same as the pre-training phase, we trained the video tokenizer on WebViD for $10$ epochs with initial learning rate $2e-4$. The embedding before/after look-up have all been $l2$-normalized for Euclidean distance computation. We further discuss the effect of tokenizer in the supplementary materials.

\subsection{Results} 

\vspace{0.5ex} \noindent\textbf{Video Question Answering.}
We compare \evlm with previous state-of-the-art methods on Video Question Answering (VQA) task on multiple datasets and present the results in Table~\ref{table:vqa}. We clearly observe that \evlm
reach on par performances or even surpasses most prior arts, where they unanimously employ large architectures. In particular, \evlm achieves $39.3\%$ Top-1 Acc. on MSRVTT-QA task, with a significant gain of $2.5\%$ Acc. over all-in-one-T~\cite{wang2022all}, which is a tiny VL architecture. Remarkable improvements are observed on T-GIF and MSVD benchmarks consistently: \evlm exceeds all-in-one-T for $+9\%$ Top-1 Acc. on MSVD, $+4.1\%$ on TGIF-Action, $+10.7\%$ on TGIF-Transition and $+6.8\%$ higher on TGIF-Frame. MC denotes the multiple choice task and \evlm also performs prominently better than its counterpart.
Notably, we note that \evlm even exceed most large VL architectures, \ie, ClipBERT~\cite{lei2021less}, CoMVT~\cite{Peng2022aunified} and SSML~\cite{amrani2021noise}, indicating the superiority of our efficient architecture which has an obvious advantage regarding the efficiency and inference speed. Also, \evlm is surprisingly more data efficient than the competitors and only exploits $2.5$M video-text pairs, whereas the baseline approaches use much larger scale pre-training: for instance, JustAsk~\cite{yang2021just} adopts $69$M video-question-answer triplets auto-extracted from narrated videos for training.; All-in-one-T has been pre-trained on much larger VL corpus, WebViD2.5M and HowTo100M~\cite{miech2019howto100m}, which consists of over $100$ million video clip-text pairs. 

\begin{table}[t]{
    \centering{
    \setlength\extrarowheight{2.6pt}
    \resizebox{.99\columnwidth}{!}
    {%
    \begin{tabular}{lcccc}
    \toprule
    \multirow{2}{*}{\textbf{Methods}}  & \multicolumn{2}{c}{\texttt{\textbf{MSRVTT}}}  & & \multicolumn{1}{c}{\texttt{\textbf{DiDeMo}}} \\ \cmidrule{2-3} \cmidrule{5-5}
    & {MC} & {text-to-video} & & {text-to-video}  \\
    \hline 
    HT100M & $52.9$ & \ \ \ $7.5$ / $21.2$ / $29.6$ & & - \\ 
    TACo & - & \ \ \ $9.8$ / $25.0$ / $33.4$ & & - \\
    VideoCLIP & - &  \ $10.4$ / $22.2$ / $30.0$ & & $16.6$ / $46.9$ / \ \ - \ \ \\
    all-in-one-B & $78.1$ & - & & - \\
    \hline
    all-in-one-T &  $59.0$ & $ \ \ \ \ \ 3.8$ /\ \ \ $9.7$ / $16.0^*$ & & \ \ \ \ \ $5.6$ / $13.2$ / $20.5^*$ \\ 
   \evlm  &  $65.2$ & \ \ \ $6.6$ / $18.2$ / $26.9$ & & \ \ \ $8.8$ / $23.0$ / $32.6$ \\
    \bottomrule
    \end{tabular}
    }
    }
    }
    \vspace{1mm}
    \caption{Zero-shot performances of \evlm on text-to-video retrieval and multiple choice (MC) tasks. We report Top-1 Acc. on MC task and recall scores for retrieval task.}
    \vspace{-2mm}
    \label{tab:zeroshot}
\end{table}

\vspace{0.5ex} \noindent\textbf{Text-to-Video Retrieval.}
We study to extend \evlm for video retrieval from text task. Table~\ref{table:retrieval} summarizes results on MSRVTT~\cite{xu2016msr} and DiDeMo~\cite{anne2017localizing} for comparisons with previous methods. We follow~\cite{wang2022all,wang2021ufo} to speed up retrieval efficiency by replacing the one-to-one image-text matching schema with encoding unimodal inputs during fine-tuning and inference. This additionally accelerates the notoriously slow retrieval process. For instance, our model only takes up $2$ minutes for one round of text-to-video retrieval on MSRVTT testing split with one V100 GPU device,
while regular VL models, however, usually take up to several hours~\cite{wang2022all} or more. We explain more about how we conduct video retrieval in the Appendix.
\evlm achieves significant performance gain over a tiny VL variation of all-in-one~\cite{wang2022all}, pre-trained across the aggregation of HowTo100M and WebViD2.5M datasets. Although \evlm leverages only $16$M parameters, it still surpasses some prevailing retrieval architectures, say, HT100M~\cite{miech2019howto100m}, HERO~\cite{li2020hero}, ClipBert~\cite{lei2021less}, and further reaches performance close to  TACo~\cite{yang2021taco}. To be specific, \evlm achieves R@1/ R@5/ R@10 recall $27.0$ / $51.7$ / $64.4$ on MSRVTT dataset, and $23.8$ / $44.9$ / $53.2$ on DiDeMo. Admittedly, \evlm still shows an obvious performance gap with state-of-the-art architectures, which majorly benefited from dozens times more parameters~\cite{wang2022all,fu2021violet,bain2021frozen}; specific optimization just for video retrieval task (Frozen~\cite{bain2021frozen}); and larger VL corpus for pre-training. Compared with them, \evlm achieves great speed \& performance trade-off and is also notably more efficient. 


\begin{table}[t]{
    \centering
    \scalebox{.88}{
    \begin{tabular}{lccccccc}
    \toprule
    \textbf{VTM}  & \textbf{MLM}  & \textbf{MVM} & & {text-to-video} & QA & MC  \\ 
    \hline
    - & - & -  & & $14.6$ / $34.8$ / $51.0$ & $32.8$ & $62.0$\\
    \checkmark & - & -  &  & $22.9$ / $45.9$ / $58.9$ & $34.5$ & $69.4$\\
    \checkmark & \checkmark & -  &  & $23.6$ / $47.6$ / $60.1$ & $35.6$ & $72.8$\\
    \checkmark & \checkmark & \checkmark  & &  $27.0$ / $51.7$ / $64.4$ & $39.3$ & $89.2$ \\
    \bottomrule
    \end{tabular}
    }
    \vspace{2mm}
    \caption{Ablative studies for effects of different modeling items on text-to-video retrieval task. First row shows the results obtained w.o. any pre-training (downstream task fine-tuning only).}
    \label{tab:ablation}
    }
\end{table}

\vspace{0.5ex} \noindent\textbf{Transferring to Activity Recognition.} In order to comprehensively study the transferability of \evlm\!, we extend it to adapt \evlm for activity recognition tasks in a linear probing manner (only the classification layer is trainable).  We choose HMDB51~\cite{kuehne2011hmdb} and Kinetics-400 datasets as our testbeds. HMDB51 contains $6,849$ video clips extracted from commercial movies, divided into $51$ human action categories, each containing a minimum of 101 clips. Kinetics-400~\cite{kay2017kinetics} dataset is a collection of $650,000$ video clips that cover $400$ human action classes. We conduct the classification training in a linear probing manner and report the Top-1 Acc. in Table~\ref{table:activity}. We observe that \evlm evidently outperforms all-in-one-T by a large margin: $41.6\%$ \vs $18.4$\%. \evlm$^\text{MVM}$ denotes the pre-trained \evlm optimized with our masked video modeling. We observe that adopting MVM further improves the Top-1 Acc. by additional $2.0\%$. This verifies that the VL representations learned have great generalization ability, and the proposed $\mathcal{L}_\text{MVM}$ further strengthens that.

\vspace{0.5ex} \noindent\textbf{Zero-shot Multiple Choice and Video Retrieval.} We now examine \evlm from the perspective of zero-shot recognition and retrieval. We report performances \evlm on both the MSRVTT and DiDeMo datasets, where the model is pre-trained and evaluated without any further fine-tuning. Table~\ref{tab:zeroshot} shows the results of \evlm with previous SOTA methods. \evlm reaches $65.2\%$ Top-1 accuracy on MSRVTT and $6.6 / 18.2 / 26.9$ (R@1/R@5/R@10) for text-to-retrieval task, which is close to such larger VL architectures as VideoCLIP and TaCo. In comparison with all-in-one-T~\cite{wang2022all}, our method noticeably exceeds it. This clearly verifies the great generalization ability of our proposed method.

\begin{table}[t]
\centering
\setlength{\tabcolsep}{1.8pt} 
\renewcommand{\arraystretch}{1.11} 
\scalebox{1.22}{
\scriptsize
\begin{tabular}{lccc}
\toprule
\textbf{Toeknizer} & VOC. RATE & {HMDB51} & {MSRVTT-QA} \\
\hline 
{Baseline} & { -} & {$33.9$} &
{$37.9$}  \\
\evlm + T$_\text{DALL-E}$ & - & $34.2$ & $38.1$ \\ 
\evlm+ T$_\text{BEVT/PECO}$ & - & $34.5$ & $38.3$ \\ 
\evlm+ T$_\text{-}$$\mathcal{L}_\text{BCE}$ & $64.2\%$ & $35.9$ & $38.7$ \\ 
\evlm+ T$_\text{-}$$\mathcal{L}_\text{TOKEN}$ & $92.0\%$ & $41.6$ & $39.3$ \\ 
\bottomrule
\end{tabular}
}
\vspace{1mm}
\caption{\scriptsize Performances of \evlm with different tokenizers from~\cite{ramesh2021zero,dong2021peco,wang2022bevt} or obtained by different strategies. 
VOC. RATE denotes code-book utilization rate and higher value indicates a broader semantic space coverage.
}
\label{tab:tokenizer}
\vspace{-4mm}
\end{table} 

\vspace{0.5ex} \noindent\textbf{Effect of MVM in VL Pre-training.} In the furtherance of gaining more understanding of the impact of MVM throughout the different stages, we illustrate their performances on MSRVTT retrieval task in Table~\ref{tab:ablation} and then compare the results w/w.o. MVM on HMDB and MSRVTT-MC tasks. We observe from the table that \evlm with our proposed MVM clearly outperforms the counterpart (without MVM) and the combination of all modeling items leads to the best result. Also, we observe that the proposed MVM modeling is complementary to MLM and VTM across multiple tasks consistently.  Tab~\ref{tab:tokenizer} shows our comparison results leveraging different video tokenizers. We compare our optimized tokenizer to~\cite{liu2021swin} (motion-level reconstruction), DALL-E (pixel-level reconstruction as~\cite{fu2021violet}), and tokenizer optimized with BCE as reconstruction loss. Our focal-loss optimized tokenizer outperforms others on QA task ($39.3$ v.s. $38.5$). \evlm\! outperforms DALL-E's pixel-level tokenizer to a large extent, confirming that semantically discretized labels are important for VL learning on small VL architectures. 
The table also shows the performance of~\cite{dong2021peco,wang2022bevt}'s official tokenizer and we observe a big performance gap ($34.5$ v.s. $41.6$) with \evlm mainly because of the lack of semantic optimization as \evlm\!. Different MVM target ($\mathcal{L}_\text{BCE}$) suffers from imbalanced words distribution and yields a less vocabulary utilization rate and performances.

\subsection{Choice of Video Tokenizer}
Performances of \evlm can be dependent on the choice of different video tokenizers. In this section, we continue to discuss the effects that Video Tokenizer has on VL pre-training. We instantiate three variations of Video Tokenizer with different visual backbones for \evlm and compare the effect of them for VL pre-training: 1). The Tokenizer from DALL-E~\cite{ramesh2021zero}, a text-to-image generation model, 2). A video SWIN based tokenizer pre-trained from the Kinetic Activity Dataset~\cite{kay2017kinetics}; and 3). A ViT~\cite{dosovitskiy2020image} based tokenizer was pre-trained on a image-text contrastive pre-training~\cite{radford2021learning}. We introduce the details of them:

\noindent \textbf{Pixel-reconstruction Tokenizer.}  Some previous works~\cite{bao2021beit,fu2021violet} leverage Tokenizer in DALL-E~\cite{ramesh2021zero} as the image tokenizer learned by discrete variational autoencoder (dVAE). A recent study proposes to train an advanced version of tokenizer~\cite{peng2022beit} providing better discrete labels for self-supervised representation learning. Overall, there are two modules during visual token learning, namely, a tokenizer and a decoder (disposed after learning). The tokenizer maps image pixels into discrete tokens via nearest neighbor search according to a learnable visual codebook (or vocabulary). And the decoder learns to reconstruct the input raw image based on the visual tokens. The major aspect that differentiates the tokenizer of \evlm is that it tries to reconstruct the semantic words instead of pixels. Tokenizer of DALL-E~\cite{ramesh2021zero} is utilized in concurrent works for self-supervised visual representation learning like~\cite{peng2022beit}, also Vision-Language representation learning~\cite{fu2022empirical}. 

\noindent \textbf{Semantic-reconstruction Tokenizer.} We propose to learn the tokenizer by only reconstructing the semantic information of the image without resorting to the RGB pixels. In our experiments, this is achieved by our proposed caption word classification, as previously described.  We construct two baselines here for comparison: 1). a Video SWIN~\cite{liu2021swin} backbone-based tokenizer; and 2). a ViT~\cite{dosovitskiy2020image} based tokenizer that has been pre-trained by~\cite{radford2021learning}, and leverages $400$M image-text pairs for contrastive pre-training. 

For both options, we adopt a similar set of hyper-parameters and settings. In specific, we use $9,420$ prototype codes, with each in $32$ dimensions. For decoders, in tokenizer's training, a $8$ layer transformer blocks are used.  Another challenge that prevents Video SWIN for our $L_\text{MVM}$ comes from its inconsistent number of output tokens due to the pooling operation along the temporal axis. And following experiences from Fu et al.,~\cite{fu2022empirical, fu2021violet}, we remove the 3D pooling layer in the architecture, and this makes the output tokens consistent with our produced tokens. Another notable disadvantage of leveraging Video SWIN comes from its large computation consumption, which largely slows the VL pre-training more than vanilla ViT. In Table~\ref{tab:tokenizer}, we report the performances of leveraging different Tokenizers, as well as the vocabulary usage rate. We observe that the ViT-based tokenizer reaches the highest performance on the VQA task, and the ViT-based method reaches a much higher vocabulary usage rate, which validates the diversity of code embedding, all contributing to the semantic reconstruction collaboratively. Interestingly, we observe that the performance gains from Video SWIN are extremely high for activity recognition tasks, yet the improvement is not that obvious for general VL tasks like VQA. We conjecture that this is caused by the pre-training of Video SWIN on Kinetics datasets~\cite{kay2017kinetics}, which is a classification task for activity recognition. Comparing with the tokenizer of DALL-E, our semantic tokenizers bring obvious better performances, though we leverage much fewer video-text pairs for pre-training. It is worth mentioning that the vocabulary rate for Video SWIN is particularly lower than our method, indicating that it may not generalize well to diverse scenes.

\begin{table}[h]
\centering
\setlength{\tabcolsep}{4pt} 
\renewcommand{\arraystretch}{1.15} 
\small
\begin{tabular}{lccc}
\toprule
& &{{ {\textbf{HMDB51}}}} & {{ {\textbf{MSRVTT-QA}}}}\\ 
& \text{VOC. RATE} & {Acc.} & {Acc.} \\
\hline 
{ Baseline} & - & {$33.9$} &
{ $37.9$}  \\
\  \evlm$_\text{+DALL-E}$ & - & $34.2$ & $38.1$ \\ 
\  \evlm$_\text{+VideoSWIN}$ & $38.9\%$ & $51.3$ & $38.5$ \\ 
\  \evlm$_\text{+ViT}$ & $92.0\%$ & $41.6$ & $39.3$ \\ 

\bottomrule
\end{tabular}
\vspace{2mm}
\caption{\small Performances of \evlm with different tokenizers. VOC. RATE denotes vocabulary usage rate. The vocabulary rate for DALL-E is not officially reported, so we neglect it.
}
\label{tab:tokenizer}
\end{table} 

\subsection{About Fast Video-Text Retrieval}
As discussed in~\cite{wang2022all,lei2022loopitr}, cross-modal architecture takes paired image/video-text pairs for matching scores, which are used for rankings in text-to-image/video retrieval. Yet, such joint retrieval can largely impede the retrieval evaluation (e.g., for 1K video-caption pairs, it yields 1K$\times$1K$=$1M iterations of forward). To circumvent this, we follow fast retrieval, as in~\cite{wang2022all}, where during downstream fine-tuning we feed in either only texts or videos as input. This, according to our empirical results, reduces the performance to some extent but largely speeds up the evaluation. Though we did not apply these text/video-only strategies during pre-training as in~\cite{wang2021ufo}, we believe our retrieval results can be further improved by sacrificing the inference efficiency or adopting a more consistent pre-training strategy, which is, however, beyond the scope of this work. This also somehow explains the limited performance of \evlm on zero-shot text-to-video retrieval tasks, where no fine-tuning has been conducted and the model naturally does not generalize well to text-only or video-only input. For the retrieval task, we fine-tune the model for 200 epochs with an initial learning rate of $3e-4$.

\subsection{Parameters of the Model }
\begin{figure}
    \centering
    \includegraphics[width=.45\textwidth]{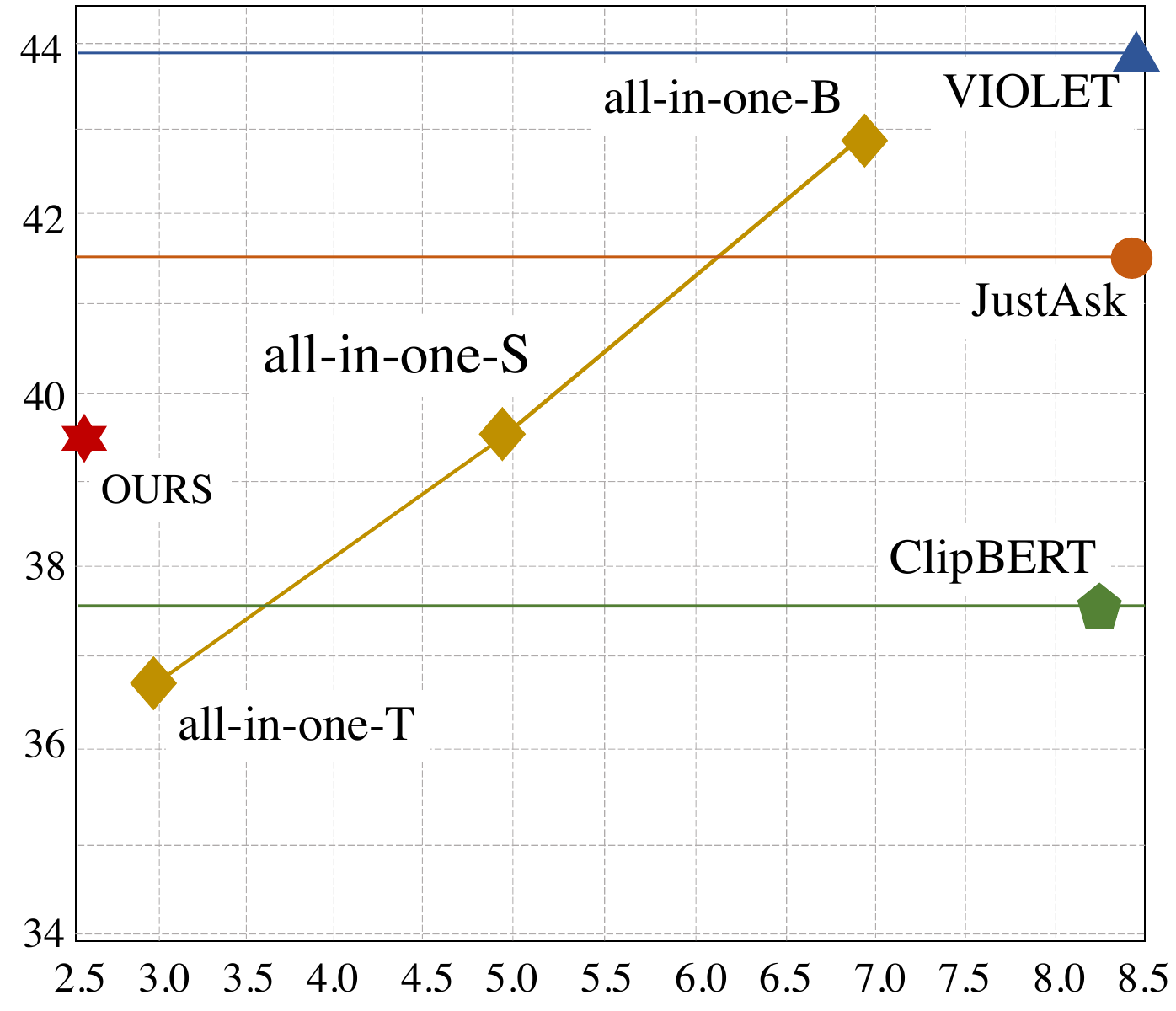}
    \vspace{-2mm}
    \caption{Comparisons of \evlm with state-of-the-art VL architectures on MSRVTT-QA. The X axis shows the values of \texttt{log(parameters)} to adapt scales.}
    \label{fig:flops}
\end{figure}

We provide the detailed breakdown of parameters in \evlm, and present it in Table~\ref{tab:param}. In Figure~\ref{fig:flops}, we compare the performances \& parameters of \evlm with state-of-the-art VL architectures on the MSRVTT QA task. We see from the picture that \evlm achieve great performance \& inference speed trade off over other competent competitors like all-in-one-T, CLIPBert.

\begin{table}[t]
\centering
\setlength{\tabcolsep}{8pt} 
\renewcommand{\arraystretch}{1.15} 
\scalebox{1.19}{
\small
\begin{tabular}{lc}
\toprule
\textbf{Layer Name} & \textbf{Params} \\
\hline 
Textual Embedding Layer & $1.2$M\\ 
Textual Encoder & $2.4$M\\ 
Textual MLP Layer & $49.5$K\\ 
Token Type Emb. Layer & $1.2$K\\ 
Visual Encoder & $5.3$M\\ 
Visual MLP Layer & $117$K\\ 
Multi-Modal Decoder & $7.1$M\\ 
Multi-modal Pooling Layer & $147$K\\ 
CLS HEAD (VQA) & $770$\\ 
\hline
TOTAL & $16.3$M\\ 
\bottomrule
\end{tabular}
}
\vspace{2mm}
\caption{\small Parameter breakdown of \evlm for VQA task on MSRVTT-QA.
}
\label{tab:param}
\end{table}

\section{Discussions}
\begin{figure}
    \centering
  \includegraphics[width=.48\textwidth]{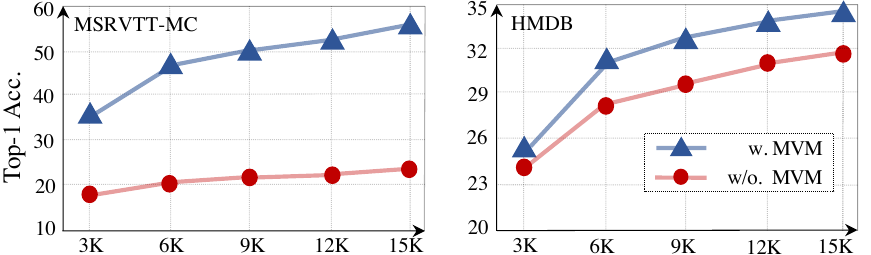}%
  \caption{Performances of \evlm with or without MVM throughout the $15$K-step VL pre-training. Left: zero-shot multiple-choice task on MSRVTT. Right: Activity recognition task on HMDB51 in the linear probing manner. }
  \label{fig:ablative}
\end{figure}

We then proceed to assess the generalization ability of intermediate-learned representations during the pre-training. As shown in Figure~\ref{fig:ablative}, we see that from training steps <$3$K, $6$K, $9$K, $12$K, $15$K>, MVM consistently outperforms the regular VL modelings to a non-trivial gap throughout the VL pre-training. This proves that our MVM brings continuous benefits during the VL pre-training and is also more data-efficient. We provide more comparisons, analytical results, and further discussion of the Video Tokenizer in the Appendix.

\section{Conclusion}
In this paper, we propose the \evlm\!, a small Video-Language model that learns cross-modal representations efficiently and generalizes well to a series of VL downstream tasks. Compared with existing Video-Language models, \evlm vastly increases the inference speed by over $40$ times with only $15\%$ of the parameters. With our proposed semantic vector-quantized video tokenizer, our VL pre-training is formulated as reconstructing the masked video region's discrete labels as well as the masked language and video-text matching modeling. Extensive experiments indicate that \evlm achieves competing performances, approaching most large Video-Language models and significantly outperforming its small counterpart. We anticipate that \evlm will lead to more future works in building efficient Vision and Language models for broader real-world impact.

{\small
\bibliographystyle{ieee_fullname}
\bibliography{egbib}

\begin{thebibliography}{10}\itemsep=-1pt

\bibitem{amrani2021noise}
Elad Amrani, Rami Ben-Ari, Daniel Rotman, and Alex Bronstein.
\newblock Noise estimation using density estimation for self-supervised multimodal learning.
\newblock In {\em Proceedings of the AAAI Conference on Artificial Intelligence}, volume~35, pages 6644--6652, 2021.

\bibitem{anderson2018bottom}
Peter Anderson, Xiaodong He, Chris Buehler, Damien Teney, Mark Johnson, Stephen Gould, and Lei Zhang.
\newblock Bottom-up and top-down attention for image captioning and visual question answering.
\newblock In {\em Proceedings of the IEEE conference on computer vision and pattern recognition}, pages 6077--6086, 2018.

\bibitem{anne2017localizing}
Lisa Anne~Hendricks, Oliver Wang, Eli Shechtman, Josef Sivic, Trevor Darrell, and Bryan Russell.
\newblock Localizing moments in video with natural language.
\newblock In {\em Proceedings of the IEEE international conference on computer vision}, pages 5803--5812, 2017.

\bibitem{mtmae}
Roman Bachmann, David Mizrahi, Andrei Atanov, and Amir Zamir.
\newblock Multimae: Multi-modal multi-task masked autoencoders, 2022.

\bibitem{bain2021frozen}
Max Bain, Arsha Nagrani, G{\"u}l Varol, and Andrew Zisserman.
\newblock Frozen in time: A joint video and image encoder for end-to-end retrieval.
\newblock In {\em Proceedings of the IEEE/CVF International Conference on Computer Vision}, pages 1728--1738, 2021.

\bibitem{bao2021beit}
Hangbo Bao, Li Dong, and Furu Wei.
\newblock Beit: Bert pre-training of image transformers.
\newblock {\em arXiv preprint arXiv:2106.08254}, 2021.

\bibitem{ben2020asymmetric}
Emanuel Ben-Baruch, Tal Ridnik, Nadav Zamir, Asaf Noy, Itamar Friedman, Matan Protter, and Lihi Zelnik-Manor.
\newblock Asymmetric loss for multi-label classification.
\newblock {\em arXiv preprint arXiv:2009.14119}, 2020.

\bibitem{bengio2013estimating}
Yoshua Bengio, Nicholas L{\'e}onard, and Aaron Courville.
\newblock Estimating or propagating gradients through stochastic neurons for conditional computation.
\newblock {\em arXiv preprint arXiv:1308.3432}, 2013.

\bibitem{brown2020language}
Tom Brown, Benjamin Mann, Nick Ryder, Melanie Subbiah, Jared~D Kaplan, Prafulla Dhariwal, Arvind Neelakantan, Pranav Shyam, Girish Sastry, Amanda Askell, et~al.
\newblock Language models are few-shot learners.
\newblock {\em Advances in neural information processing systems}, 33:1877--1901, 2020.

\bibitem{cao2020behind}
Jize. Cao.
\newblock Behind the scene: Revealing the secrets of pre-trained vision-and-language models.
\newblock In {\em European Conference on Computer Vision.}, 2020.

\bibitem{cao2022mobivqa}
Qingqing Cao, Prerna Khanna, Nicholas~D Lane, and Aruna Balasubramanian.
\newblock Mobivqa: Efficient on-device visual question answering.
\newblock 2022.

\bibitem{chen2011collecting}
David Chen and William~B Dolan.
\newblock Collecting highly parallel data for paraphrase evaluation.
\newblock In {\em Proceedings of the 49th annual meeting of the association for computational linguistics: human language technologies}, pages 190--200, 2011.

\bibitem{chen2019uniter}
Yen-Chun Chen, Linjie Li, Licheng Yu, Ahmed El~Kholy, Faisal Ahmed, Zhe Gan, Yu Cheng, and Jingjing Liu.
\newblock Uniter: Learning universal image-text representations.
\newblock 2019.

\bibitem{devlin2018bert}
Jacob Devlin, Ming-Wei Chang, Kenton Lee, and Kristina Toutanova.
\newblock Bert: Pre-training of deep bidirectional transformers for language understanding.
\newblock {\em arXiv preprint arXiv:1810.04805}, 2018.

\bibitem{dong2021peco}
Xiaoyi Dong, Jianmin Bao, Ting Zhang, Dongdong Chen, Weiming Zhang, Lu Yuan, Dong Chen, Fang Wen, and Nenghai Yu.
\newblock Peco: Perceptual codebook for bert pre-training of vision transformers.
\newblock {\em arXiv preprint arXiv:2111.12710}, 2021.

\bibitem{dosovitskiy2020image}
Alexey Dosovitskiy, Lucas Beyer, Alexander Kolesnikov, Dirk Weissenborn, Xiaohua Zhai, Thomas Unterthiner, Mostafa Dehghani, Matthias Minderer, Georg Heigold, Sylvain Gelly, et~al.
\newblock An image is worth 16x16 words: Transformers for image recognition at scale.
\newblock {\em arXiv preprint arXiv:2010.11929}, 2020.

\bibitem{dou2022empirical}
Zi-Yi Dou, Yichong Xu, Zhe Gan, Jianfeng Wang, Shuohang Wang, Lijuan Wang, Chenguang Zhu, Pengchuan Zhang, Lu Yuan, Nanyun Peng, et~al.
\newblock An empirical study of training end-to-end vision-and-language transformers.
\newblock In {\em Proceedings of the IEEE/CVF Conference on Computer Vision and Pattern Recognition}, pages 18166--18176, 2022.

\bibitem{esser2021taming}
Patrick Esser, Robin Rombach, and Bjorn Ommer.
\newblock Taming transformers for high-resolution image synthesis.
\newblock In {\em Proceedings of the IEEE/CVF conference on computer vision and pattern recognition}, pages 12873--12883, 2021.

\bibitem{fang2022injecting}
Zhiyuan Fang, Jianfeng Wang, Xiaowei Hu, Lin Liang, Zhe Gan, Lijuan Wang, Yezhou Yang, and Zicheng Liu.
\newblock Injecting semantic concepts into end-to-end image captioning.
\newblock In {\em Proceedings of the IEEE/CVF Conference on Computer Vision and Pattern Recognition}, pages 18009--18019, 2022.

\bibitem{fang2021compressing}
Zhiyuan Fang, Jianfeng Wang, Xiaowei Hu, Lijuan Wang, Yezhou Yang, and Zicheng Liu.
\newblock Compressing visual-linguistic model via knowledge distillation.
\newblock In {\em Proceedings of the IEEE/CVF International Conference on Computer Vision}, pages 1428--1438, 2021.

\bibitem{feichtenhofer2022masked}
Christoph Feichtenhofer, Haoqi Fan, Yanghao Li, and Kaiming He.
\newblock Masked autoencoders as spatiotemporal learners.
\newblock {\em arXiv preprint arXiv:2205.09113}, 2022.

\bibitem{videoMAE2}
Christoph Feichtenhofer, Haoqi Fan, Yanghao Li, and Kaiming He.
\newblock Masked autoencoders as spatiotemporal learners, 2022.

\bibitem{fu2021violet}
Tsu-Jui Fu, Linjie Li, Zhe Gan, Kevin Lin, William~Yang Wang, Lijuan Wang, and Zicheng Liu.
\newblock Violet: End-to-end video-language transformers with masked visual-token modeling.
\newblock {\em arXiv preprint arXiv:2111.12681}, 2021.

\bibitem{fu2022empirical}
Tsu-Jui Fu, Linjie Li, Zhe Gan, Kevin Lin, William~Yang Wang, Lijuan Wang, and Zicheng Liu.
\newblock An empirical study of end-to-end video-language transformers with masked visual modeling.
\newblock {\em arXiv preprint arXiv:2209.01540}, 2022.

\bibitem{gao2022transform}
Feng Gao, Qing Ping, Govind Thattai, Aishwarya Reganti, Ying~Nian Wu, and Prem Natarajan.
\newblock Transform-retrieve-generate: Natural language-centric outside-knowledge visual question answering.
\newblock In {\em Proceedings of the IEEE/CVF Conference on Computer Vision and Pattern Recognition}, pages 5067--5077, 2022.

\bibitem{geng2022multimodal}
Xinyang Geng, Hao Liu, Lisa Lee, Dale Schuurams, Sergey Levine, and Pieter Abbeel.
\newblock Multimodal masked autoencoders learn transferable representations.
\newblock {\em arXiv preprint arXiv:2205.14204}, 2022.

\bibitem{goel2022cyclip}
Shashank Goel, Hritik Bansal, Sumit Bhatia, Ryan~A Rossi, Vishwa Vinay, and Aditya Grover.
\newblock Cyclip: Cyclic contrastive language-image pretraining.
\newblock {\em arXiv preprint arXiv:2205.14459}, 2022.

\bibitem{he2022masked}
Kaiming He, Xinlei Chen, Saining Xie, Yanghao Li, Piotr Doll{\'a}r, and Ross Girshick.
\newblock Masked autoencoders are scalable vision learners.
\newblock In {\em Proceedings of the IEEE/CVF Conference on Computer Vision and Pattern Recognition}, pages 16000--16009, 2022.

\bibitem{huo2021wenlan}
Yuqi Huo, Manli Zhang, Guangzhen Liu, Haoyu Lu, Yizhao Gao, Guoxing Yang, Jingyuan Wen, Heng Zhang, Baogui Xu, Weihao Zheng, et~al.
\newblock Wenlan: Bridging vision and language by large-scale multi-modal pre-training.
\newblock {\em arXiv preprint arXiv:2103.06561}, 2021.

\bibitem{perceiverIO}
Andrew Jaegle, Sebastian Borgeaud, Jean-Baptiste Alayrac, Carl Doersch, Catalin Ionescu, David Ding, Skanda Koppula, Daniel Zoran, Andrew Brock, Evan Shelhamer, Olivier Hénaff, Matthew~M. Botvinick, Andrew Zisserman, Oriol Vinyals, and Joāo Carreira.
\newblock Perceiver io: A general architecture for structured inputs amp; outputs, 2021.

\bibitem{perceiver}
Andrew Jaegle, Felix Gimeno, Andrew Brock, Andrew Zisserman, Oriol Vinyals, and Joao Carreira.
\newblock Perceiver: General perception with iterative attention, 2021.

\bibitem{jang2017tgif}
Yunseok Jang, Yale Song, Youngjae Yu, Youngjin Kim, and Gunhee Kim.
\newblock Tgif-qa: Toward spatio-temporal reasoning in visual question answering.
\newblock In {\em Proceedings of the IEEE conference on computer vision and pattern recognition}, pages 2758--2766, 2017.

\bibitem{jia2021scaling}
Chao Jia, Yinfei Yang, Ye Xia, Yi-Ting Chen, Zarana Parekh, Hieu Pham, Quoc Le, Yun-Hsuan Sung, Zhen Li, and Tom Duerig.
\newblock Scaling up visual and vision-language representation learning with noisy text supervision.
\newblock In {\em International Conference on Machine Learning}, pages 4904--4916. PMLR, 2021.

\bibitem{jiang2020defense}
Huaizu Jiang, Ishan Misra, Marcus Rohrbach, Erik Learned-Miller, and Xinlei Chen.
\newblock In defense of grid features for visual question answering.
\newblock In {\em Proceedings of the IEEE/CVF Conference on Computer Vision and Pattern Recognition}, pages 10267--10276, 2020.

\bibitem{jiao2019tinybert}
Xiaoqi Jiao, Yichun Yin, Lifeng Shang, Xin Jiang, Xiao Chen, Linlin Li, Fang Wang, and Qun Liu.
\newblock Tinybert: Distilling bert for natural language understanding.
\newblock {\em arXiv preprint arXiv:1909.10351}, 2019.

\bibitem{jinpeng2022all}
Alex Jinpeng~Wang, Yixiao Ge, Rui Yan, Yuying Ge, Xudong Lin, Guanyu Cai, Jianping Wu, Ying Shan, Xiaohu Qie, and Mike~Zheng Shou.
\newblock All in one: Exploring unified video-language pre-training.
\newblock {\em arXiv e-prints}, pages arXiv--2203, 2022.

\bibitem{kay2017kinetics}
Will Kay, Joao Carreira, Karen Simonyan, Brian Zhang, Chloe Hillier, Sudheendra Vijayanarasimhan, Fabio Viola, Tim Green, Trevor Back, Paul Natsev, et~al.
\newblock The kinetics human action video dataset.
\newblock {\em arXiv preprint arXiv:1705.06950}, 2017.

\bibitem{kim2021vilt}
Wonjae Kim, Bokyung Son, and Ildoo Kim.
\newblock Vilt: Vision-and-language transformer without convolution or region supervision.
\newblock In {\em International Conference on Machine Learning}, pages 5583--5594. PMLR, 2021.

\bibitem{krishna2017visual}
Ranjay Krishna, Yuke Zhu, Oliver Groth, Justin Johnson, Kenji Hata, Joshua Kravitz, Stephanie Chen, Yannis Kalantidis, Li-Jia Li, David~A Shamma, et~al.
\newblock Visual genome: Connecting language and vision using crowdsourced dense image annotations.
\newblock {\em International journal of computer vision}, 123(1):32--73, 2017.

\bibitem{kuehne2011hmdb}
Hildegard Kuehne, Hueihan Jhuang, Est{\'\i}baliz Garrote, Tomaso Poggio, and Thomas Serre.
\newblock Hmdb: a large video database for human motion recognition.
\newblock In {\em 2011 International conference on computer vision}, pages 2556--2563. IEEE, 2011.

\bibitem{lei2022loopitr}
Jie Lei, Xinlei Chen, Ning Zhang, Mengjiao Wang, Mohit Bansal, Tamara~L Berg, and Licheng Yu.
\newblock Loopitr: Combining dual and cross encoder architectures for image-text retrieval.
\newblock {\em arXiv preprint arXiv:2203.05465}, 2022.

\bibitem{lei2021less}
Jie Lei, Linjie Li, Luowei Zhou, Zhe Gan, Tamara~L Berg, Mohit Bansal, and Jingjing Liu.
\newblock Less is more: Clipbert for video-and-language learning via sparse sampling.
\newblock In {\em Proceedings of the IEEE/CVF Conference on Computer Vision and Pattern Recognition}, pages 7331--7341, 2021.

\bibitem{li2020unicoder}
Gen Li, Nan Duan, Yuejian Fang, Ming Gong, and Daxin Jiang.
\newblock Unicoder-vl: A universal encoder for vision and language by cross-modal pre-training.
\newblock In {\em Proceedings of the AAAI Conference on Artificial Intelligence}, volume~34, pages 11336--11344, 2020.

\bibitem{li2020hero}
Linjie Li, Yen-Chun Chen, Yu Cheng, Zhe Gan, Licheng Yu, and Jingjing Liu.
\newblock Hero: Hierarchical encoder for video+ language omni-representation pre-training.
\newblock {\em arXiv preprint arXiv:2005.00200}, 2020.

\bibitem{li2022uniform}
Xiang Li, Wenhai Wang, Lingfeng Yang, and Jian Yang.
\newblock Uniform masking: Enabling mae pre-training for pyramid-based vision transformers with locality.
\newblock {\em arXiv preprint arXiv:2205.10063}, 2022.

\bibitem{li2020oscar}
Xiujun Li, Xi Yin, Chunyuan Li, Pengchuan Zhang, Xiaowei Hu, Lei Zhang, Lijuan Wang, Houdong Hu, Li Dong, Furu Wei, et~al.
\newblock Oscar: Object-semantics aligned pre-training for vision-language tasks.
\newblock In {\em European Conference on Computer Vision}, pages 121--137. Springer, 2020.

\bibitem{li2016tgif}
Yuncheng Li, Yale Song, Liangliang Cao, Joel Tetreault, Larry Goldberg, Alejandro Jaimes, and Jiebo Luo.
\newblock Tgif: A new dataset and benchmark on animated gif description.
\newblock In {\em Proceedings of the IEEE Conference on Computer Vision and Pattern Recognition}, pages 4641--4650, 2016.

\bibitem{lin2014microsoft}
Tsung-Yi Lin, Michael Maire, Serge Belongie, James Hays, Pietro Perona, Deva Ramanan, Piotr Doll{\'a}r, and C~Lawrence Zitnick.
\newblock Microsoft coco: Common objects in context.
\newblock In {\em European conference on computer vision}, pages 740--755. Springer, 2014.

\bibitem{liu2021swin}
Ze Liu, Yutong Lin, Yue Cao, Han Hu, Yixuan Wei, Zheng Zhang, Stephen Lin, and Baining Guo.
\newblock Swin transformer: Hierarchical vision transformer using shifted windows.
\newblock In {\em Proceedings of the IEEE/CVF International Conference on Computer Vision}, pages 10012--10022, 2021.

\bibitem{liu2022video}
Ze Liu, Jia Ning, Yue Cao, Yixuan Wei, Zheng Zhang, Stephen Lin, and Han Hu.
\newblock Video swin transformer.
\newblock In {\em Proceedings of the IEEE/CVF Conference on Computer Vision and Pattern Recognition}, pages 3202--3211, 2022.

\bibitem{loshchilov2017decoupled}
Ilya Loshchilov and Frank Hutter.
\newblock Decoupled weight decay regularization.
\newblock {\em arXiv preprint arXiv:1711.05101}, 2017.

\bibitem{lu2022unified}
Jiasen Lu, Christopher Clark, Rowan Zellers, Roozbeh Mottaghi, and Aniruddha Kembhavi.
\newblock Unified-io: A unified model for vision, language, and multi-modal tasks.
\newblock {\em arXiv preprint arXiv:2206.08916}, 2022.

\bibitem{luo2020univl}
Huaishao Luo, Lei Ji, Botian Shi, Haoyang Huang, Nan Duan, Tianrui Li, Jason Li, Taroon Bharti, and Ming Zhou.
\newblock Univl: A unified video and language pre-training model for multimodal understanding and generation.
\newblock {\em arXiv preprint arXiv:2002.06353}, 2020.

\bibitem{maaz2022edgenext}
Muhammad Maaz, Abdelrahman Shaker, Hisham Cholakkal, Salman Khan, Syed~Waqas Zamir, Rao~Muhammad Anwer, and Fahad~Shahbaz Khan.
\newblock Edgenext: efficiently amalgamated cnn-transformer architecture for mobile vision applications.
\newblock {\em arXiv preprint arXiv:2206.10589}, 2022.

\bibitem{miech2020end}
Antoine Miech, Jean-Baptiste Alayrac, Lucas Smaira, Ivan Laptev, Josef Sivic, and Andrew Zisserman.
\newblock End-to-end learning of visual representations from uncurated instructional videos.
\newblock In {\em Proceedings of the IEEE/CVF Conference on Computer Vision and Pattern Recognition}, pages 9879--9889, 2020.

\bibitem{miech2019howto100m}
Antoine Miech, Dimitri Zhukov, Jean-Baptiste Alayrac, Makarand Tapaswi, Ivan Laptev, and Josef Sivic.
\newblock Howto100m: Learning a text-video embedding by watching hundred million narrated video clips.
\newblock In {\em Proceedings of the IEEE/CVF International Conference on Computer Vision}, pages 2630--2640, 2019.

\bibitem{min2022voxel}
Chen Min, Dawei Zhao, Liang Xiao, Yiming Nie, and Bin Dai.
\newblock Voxel-mae: Masked autoencoders for pre-training large-scale point clouds.
\newblock {\em arXiv preprint arXiv:2206.09900}, 2022.

\bibitem{mu2022slip}
Norman Mu, Alexander Kirillov, David Wagner, and Saining Xie.
\newblock Slip: Self-supervision meets language-image pre-training.
\newblock In {\em European Conference on Computer Vision}, pages 529--544. Springer, 2022.

\bibitem{peng2022beit}
Zhiliang Peng, Li Dong, Hangbo Bao, Qixiang Ye, and Furu Wei.
\newblock Beit v2: Masked image modeling with vector-quantized visual tokenizers.
\newblock {\em arXiv preprint arXiv:2208.06366}, 2022.

\bibitem{radford2021learning}
Alec Radford, Jong~Wook Kim, Chris Hallacy, Aditya Ramesh, Gabriel Goh, Sandhini Agarwal, Girish Sastry, Amanda Askell, Pamela Mishkin, Jack Clark, et~al.
\newblock Learning transferable visual models from natural language supervision.
\newblock In {\em International Conference on Machine Learning}, pages 8748--8763. PMLR, 2021.

\bibitem{radford2018improving}
Alec Radford, Karthik Narasimhan, Tim Salimans, Ilya Sutskever, et~al.
\newblock Improving language understanding by generative pre-training.
\newblock 2018.

\bibitem{ramesh2022hierarchical}
Aditya Ramesh, Prafulla Dhariwal, Alex Nichol, Casey Chu, and Mark Chen.
\newblock Hierarchical text-conditional image generation with clip latents.
\newblock {\em arXiv preprint arXiv:2204.06125}, 2022.

\bibitem{ramesh2021zero}
Aditya Ramesh, Mikhail Pavlov, Gabriel Goh, Scott Gray, Chelsea Voss, Alec Radford, Mark Chen, and Ilya Sutskever.
\newblock Zero-shot text-to-image generation.
\newblock In {\em International Conference on Machine Learning}, pages 8821--8831. PMLR, 2021.

\bibitem{ren2015faster}
Shaoqing Ren, Kaiming He, Ross Girshick, and Jian Sun.
\newblock Faster r-cnn: Towards real-time object detection with region proposal networks.
\newblock {\em Advances in neural information processing systems}, 28, 2015.

\bibitem{seo2021look}
Paul~Hongsuck Seo, Arsha Nagrani, and Cordelia Schmid.
\newblock Look before you speak: Visually contextualized utterances.
\newblock In {\em Proceedings of the IEEE/CVF Conference on Computer Vision and Pattern Recognition}, pages 16877--16887, 2021.

\bibitem{sharma2018conceptual}
Piyush Sharma, Nan Ding, Sebastian Goodman, and Radu Soricut.
\newblock Conceptual captions: A cleaned, hypernymed, image alt-text dataset for automatic image captioning.
\newblock In {\em Proceedings of the 56th Annual Meeting of the Association for Computational Linguistics (Volume 1: Long Papers)}, pages 2556--2565, 2018.

\bibitem{shen2021efficient}
Zhuoran Shen, Mingyuan Zhang, Haiyu Zhao, Shuai Yi, and Hongsheng Li.
\newblock Efficient attention: Attention with linear complexities.
\newblock In {\em Proceedings of the IEEE/CVF winter conference on applications of computer vision}, pages 3531--3539, 2021.

\bibitem{flava}
Amanpreet Singh, Ronghang Hu, Vedanuj Goswami, Guillaume Couairon, Wojciech Galuba, Marcus Rohrbach, and Douwe Kiela.
\newblock Flava: A foundational language and vision alignment model, 2021.

\bibitem{su2019vl}
Weijie Su, Xizhou Zhu, Yue Cao, Bin Li, Lewei Lu, Furu Wei, and Jifeng Dai.
\newblock Vl-bert: Pre-training of generic visual-linguistic representations.
\newblock {\em arXiv preprint arXiv:1908.08530}, 2019.

\bibitem{sun2019videobert}
Chen Sun, Austin Myers, Carl Vondrick, Kevin Murphy, and Cordelia Schmid.
\newblock Videobert: A joint model for video and language representation learning.
\newblock In {\em Proceedings of the IEEE/CVF International Conference on Computer Vision}, pages 7464--7473, 2019.

\bibitem{tan2019efficientnet}
Mingxing Tan and Quoc Le.
\newblock Efficientnet: Rethinking model scaling for convolutional neural networks.
\newblock In {\em International conference on machine learning}, pages 6105--6114. PMLR, 2019.

\bibitem{tong2022videomae}
Zhan Tong, Yibing Song, Jue Wang, and Limin Wang.
\newblock Videomae: Masked autoencoders are data-efficient learners for self-supervised video pre-training.
\newblock {\em arXiv preprint arXiv:2203.12602}, 2022.

\bibitem{videoMAE}
Zhan Tong, Yibing Song, Jue Wang, and Limin Wang.
\newblock Videomae: Masked autoencoders are data-efficient learners for self-supervised video pre-training, 2022.

\bibitem{tsai2019small}
Henry Tsai, Jason Riesa, Melvin Johnson, Naveen Arivazhagan, Xin Li, and Amelia Archer.
\newblock Small and practical bert models for sequence labeling.
\newblock {\em arXiv preprint arXiv:1909.00100}, 2019.

\bibitem{van2017neural}
Aaron Van Den~Oord, Oriol Vinyals, et~al.
\newblock Neural discrete representation learning.
\newblock {\em Advances in neural information processing systems}, 30, 2017.

\bibitem{vaswani2017attention}
Ashish Vaswani, Noam Shazeer, Niki Parmar, Jakob Uszkoreit, Llion Jones, Aidan~N Gomez, {\L}ukasz Kaiser, and Illia Polosukhin.
\newblock Attention is all you need.
\newblock {\em Advances in neural information processing systems}, 30, 2017.

\bibitem{wang2022all}
Alex~Jinpeng Wang, Yixiao Ge, Rui Yan, Yuying Ge, Xudong Lin, Guanyu Cai, Jianping Wu, Ying Shan, Xiaohu Qie, and Mike~Zheng Shou.
\newblock All in one: Exploring unified video-language pre-training.
\newblock {\em arXiv preprint arXiv:2203.07303}, 2022.

\bibitem{wang2022object}
Jinpeng Wang, Yixiao Ge, Guanyu Cai, Rui Yan, Xudong Lin, Ying Shan, Xiaohu Qie, and Mike~Zheng Shou.
\newblock Object-aware video-language pre-training for retrieval.
\newblock In {\em Proceedings of the IEEE/CVF Conference on Computer Vision and Pattern Recognition}, pages 3313--3322, 2022.

\bibitem{wang2021ufo}
Jianfeng Wang, Xiaowei Hu, Zhe Gan, Zhengyuan Yang, Xiyang Dai, Zicheng Liu, Yumao Lu, and Lijuan Wang.
\newblock Ufo: A unified transformer for vision-language representation learning.
\newblock {\em arXiv preprint arXiv:2111.10023}, 2021.

\bibitem{wang2020minivlm}
Jianfeng Wang, Xiaowei Hu, Pengchuan Zhang, Xiujun Li, Lijuan Wang, Lei Zhang, Jianfeng Gao, and Zicheng Liu.
\newblock Minivlm: A smaller and faster vision-language model.
\newblock {\em arXiv preprint arXiv:2012.06946}, 2020.

\bibitem{wang2022git}
Jianfeng Wang, Zhengyuan Yang, Xiaowei Hu, Linjie Li, Kevin Lin, Zhe Gan, Zicheng Liu, Ce Liu, and Lijuan Wang.
\newblock Git: A generative image-to-text transformer for vision and language.
\newblock {\em arXiv preprint arXiv:2205.14100}, 2022.

\bibitem{wang2022bevt}
Rui Wang, Dongdong Chen, Zuxuan Wu, Yinpeng Chen, Xiyang Dai, Mengchen Liu, Yu-Gang Jiang, Luowei Zhou, and Lu Yuan.
\newblock Bevt: Bert pretraining of video transformers.
\newblock In {\em Proceedings of the IEEE/CVF Conference on Computer Vision and Pattern Recognition}, pages 14733--14743, 2022.

\bibitem{wang2022efficientvlm}
Tiannan Wang, Wangchunshu Zhou, Yan Zeng, and Xinsong Zhang.
\newblock Efficientvlm: Fast and accurate vision-language models via knowledge distillation and modal-adaptive pruning.
\newblock {\em arXiv preprint arXiv:2210.07795}, 2022.

\bibitem{wang2021distilled}
Zekun Wang, Wenhui Wang, Haichao Zhu, Ming Liu, Bing Qin, and Furu Wei.
\newblock Distilled dual-encoder model for vision-language understanding.
\newblock {\em arXiv preprint arXiv:2112.08723}, 2021.

\bibitem{xu2021vlm}
Hu Xu, Gargi Ghosh, Po-Yao Huang, Prahal Arora, Masoumeh Aminzadeh, Christoph Feichtenhofer, Florian Metze, and Luke Zettlemoyer.
\newblock Vlm: Task-agnostic video-language model pre-training for video understanding.
\newblock {\em arXiv preprint arXiv:2105.09996}, 2021.

\bibitem{xu2021videoclip}
Hu Xu, Gargi Ghosh, Po-Yao Huang, Dmytro Okhonko, Armen Aghajanyan, Florian Metze, Luke Zettlemoyer, and Christoph Feichtenhofer.
\newblock Videoclip: Contrastive pre-training for zero-shot video-text understanding.
\newblock {\em arXiv preprint arXiv:2109.14084}, 2021.

\bibitem{xu2016msr}
Jun Xu, Tao Mei, Ting Yao, and Yong Rui.
\newblock Msr-vtt: A large video description dataset for bridging video and language.
\newblock In {\em Proceedings of the IEEE conference on computer vision and pattern recognition}, pages 5288--5296, 2016.

\bibitem{yan2022multi}
Zhiqiang Yan, Xiang Li, Kun Wang, Zhenyu Zhang, Jun Li, and Jian Yang.
\newblock Multi-modal masked pre-training for monocular panoramic depth completion.
\newblock {\em arXiv preprint arXiv:2203.09855}, 2022.

\bibitem{yang2021just}
Antoine Yang, Antoine Miech, Josef Sivic, Ivan Laptev, and Cordelia Schmid.
\newblock Just ask: Learning to answer questions from millions of narrated videos.
\newblock In {\em Proceedings of the IEEE/CVF International Conference on Computer Vision}, pages 1686--1697, 2021.

\bibitem{yang2021taco}
Jianwei Yang, Yonatan Bisk, and Jianfeng Gao.
\newblock Taco: Token-aware cascade contrastive learning for video-text alignment.
\newblock In {\em Proceedings of the IEEE/CVF International Conference on Computer Vision}, pages 11562--11572, 2021.

\bibitem{yuan2021florence}
Lu Yuan, Dongdong Chen, Yi-Ling Chen, Noel Codella, Xiyang Dai, Jianfeng Gao, Houdong Hu, Xuedong Huang, Boxin Li, Chunyuan Li, et~al.
\newblock Florence: A new foundation model for computer vision.
\newblock {\em arXiv preprint arXiv:2111.11432}, 2021.

\bibitem{zhang2021vinvl}
Pengchuan Zhang, Xiujun Li, Xiaowei Hu, Jianwei Yang, Lei Zhang, Lijuan Wang, Yejin Choi, and Jianfeng Gao.
\newblock Vinvl: Revisiting visual representations in vision-language models.
\newblock In {\em Proceedings of the IEEE/CVF Conference on Computer Vision and Pattern Recognition}, pages 5579--5588, 2021.

\bibitem{Peng2022aunified}
Peng Zhiliang, Dong Li, Bao Hangbo, Ye Qixiang, and Wei Furu.
\newblock A unified view of masked image modeling.
\newblock {\em arXiv preprint arXiv:2210.10615}, 2022.

\end{thebibliography}
}

\end{document}